\documentclass[final,times,authoryear,twoside,american]{elsart}
\usepackage[T1]{fontenc}
\usepackage[latin1]{inputenc}
\usepackage{float}
\usepackage{graphicx}

\makeatletter

\providecommand{\tabularnewline}{\\}
\floatstyle{ruled}
\newfloat{algorithm}{tbp}{loa}
\floatname{algorithm}{Algorithm}

\usepackage{times}
\usepackage{algorithmic}
\usepackage{babel}
\makeatother
\begin{document}
\begin{frontmatter}

\title{Robust Localization and Tracking of Simultaneous Moving Sound Sources
Using Beamforming and Particle Filtering}

\author{Jean-Marc Valin$*\dagger$, Fran\c{c}ois Michaud$\dagger$, Jean
Rouat$\dagger$}

\address{$*$CSIRO ICT Centre, Marsfield, NSW 2122, Australia\\
$\dagger$Department of Electrical Engineering and Computer Engineering\\
Universit\'{e} de Sherbrooke, Sherbrooke, Quebec, J1K 2R1, Canada}

\ead[url]{http://people.xiph.org/$\sim$jm/}

\ead{\{jean-marc.valin,francois.michaud,jean.rouat\}@usherbrooke.ca}

\begin{abstract}
Mobile robots in real-life settings would benefit from being able
to localize and track sound sources. Such a capability can help localizing
a person or an interesting event in the environment, and also provides
enhanced processing for other capabilities such as speech recognition.
To give this capability to a robot, the challenge is not only to localize
simultaneous sound sources, but to track them over time. In this paper
we propose a robust sound source localization and tracking method
using an array of eight microphones. The method is based on a frequency-domain
implementation of a steered beamformer along with a particle filter-based
tracking algorithm. Results show that a mobile robot can localize
and track in real-time multiple moving sources of different types
over a range of 7 meters. These new capabilities allow a mobile robot
to interact using more natural means with people in real life settings.
\end{abstract}
\end{frontmatter}

\section{Introduction}

Sound source localization is defined as the determination of the coordinates
of sound sources in relation to a point in space. The auditory system
of living creatures provides vast amounts of information about the
world, such as localization of sound sources. For us humans, it means
to be able to focus our attention on events and changes surrounding
us, such as a cordless phone ringing, a vehicle honking, a person
who is talking to us, etc. Hearing complements well other sensors
such as vision by being omni-directional, capable to work in the dark
and not limited by physical structure (such as walls). For those who
do not have hearing impairments, it is hard to imagine going a day
without being able to hear, especially having to move in a very dynamic
and unpredictable world. Marschark \cite{Marschark} even suggests
that although deaf children have similar IQ results compared to other
children, they do experience more learning difficulties in school.
So, the intelligence manifested by autonomous robots will surely be
influenced by providing them with auditory capabilities. 

To perform sound localization, our brain combines timing (more specifically
delay or phase) and amplitude information from the sound perceived
by two ears \cite{Hartmann1999}, sometimes in addition to information
from other senses. However, localizing sound sources using only two
inputs is a challenging task. The human auditory system is very complex
and resolves the problem by accounting for the acoustic diffraction
around the head and the ridges of the outer ear. Without this ability,
localization with two microphones is limited to azimuth only, along
with the impossibility to distinguish if the sounds come from the
front or the back. Also, obtaining high-precision readings when the
sound source is in the same axis as the pair of microphones is more
difficult.

One advantage with robots is that they do not have to inherit the
same limitations as living creatures. Using more than two microphones
allows reliable and accurate localization in both azimuth and elevation.
Also, having multiple signals provides additional redundancy, reducing
the uncertainty caused by the noise and non-ideal conditions such
as reverberation and imperfect microphones. It is with this principle
in mind that we have developed an approach allowing to localize sound
sources using an array of microphones. 

Our approach is based on a frequency-domain beamformer that is steered
in all possible directions to detect sources. Instead of measuring
TDOAs and then converting to a position, the localization process
is performed in a single step. This makes the system more robust,
especially in the case where an obstacle prevents one or more microphones
from properly receiving the signals. The results of the localization
process are then enhanced by probability-based post-processing which
prevents false detection of sources. This makes the system sensitive
enough for simultaneous localization of multiple moving sound sources.
This approach is an extension of earlier work \cite{ValinICRA2004}
and works for both far-field and near-field sound sources. Detection
reliability, accuracy, and tracking capabilities of the approach are
validated using a mobile robot, with different types of sound sources.
We consider both our robust steered beamformer and our probabilistic
post-processing to contain significant contributions to the subject
of robust localization of sound sources.

The paper is organized as follows. Section \ref{sec:Related-work}
situates our work in relation to other research projects in the field.
Section \ref{sec:System-overview} presents a brief overview of the
system. Section \ref{sec:Localization-by-steered} describes our steered
beamformer implemented in the frequency-domain. Section \ref{sec:Probabilistic-post-processing}
explains how we enhance the results from the beamformer using a probabilistic
post-processor. This is followed by experimental results in Section
\ref{sec:Results}, showing how the system behaves under different
conditions. Section \ref{sec:Conclusion} concludes the paper and
presents future work.

\section{Related work\label{sec:Related-work}}

Signal processing research that addresses artificial audition is often
geared toward specific tasks such as speaker tracking for videoconferencing
\cite{Mungamuru2004}. An artificial audition system for a mobile
robot can be used for three purposes: 1) localizing sound sources;
2) separating sound sources in order to process only signals that
are relevant to a particular event in the environment; and 3) processing
sound sources to extract useful information from the environment (like
speech recognition). 

Even though artificial audition on mobile robots is a research area
still in its infancy, most of the work has been done in relation to
localization of sound sources and mostly using only two microphones.
This is the case of the SIG robot that uses both inter-aural phase
difference (IPD) and inter-aural intensity difference (IID) to locate
sounds \cite{nakadai-matsuura-okuno-kitano2003}. The binaural approach
has limitations when it comes to evaluating elevation and usually,
the front-back ambiguity cannot be resolved without resorting to active
audition \cite{Nakadai2000}.

More recently, approaches using more than two microphones have been
developed. One approach uses a circular array of eight microphones
to locate sound sources \cite{Asano2001}. In our previous work also
using eight microphones \cite{ValinIROS2003}, we presented a method
for localizing a single sound source where time delay of arrival (TDOA)
estimation was separated from the direction of arrival (DOA) estimation.
It was found that a system combining TDOA and DOA estimation in a
single step improves the system's robustness, while allowing localization
(but not tracking) of simultaneous sources \cite{ValinICRA2004}.
Kagami \emph{et al.} \cite{Kagami2004} reports a system using 128
microphones for 2D sound localization of sound sources: obviously,
it would not be practical to include such a large number of microphones
on a mobile robot.

Most of the work so far on localization of source sources does not
address the problem of tracking moving sources. It is proposed in
\cite{Bechler2004} to use a Kalman filter for tracking a moving source.
However the proposed method assumes that a single source is present.
In the past years, particle filtering \cite{Arulampalam2002} (a sequential
Monte Carlo method) has been increasingly popular to resolve object
tracking problems. Ward \emph{et al.} \cite{Ward2002b,Ward2003} and
Vermaak \cite{Vermaak2001} use this technique for tracking single
sound sources. Asoh \emph{et al.} \cite{Asoh2004} even suggested
to use this technique for mixing audio and video data to track speakers.
But again, the technique is limited to a single source due to the
problem of associating the localization observation data to each of
the sources being tracked. We refer to that problem as the source-observation
assignment problem. Some attempts are made at defining multi-modal
particle filters in \cite{Vermaak2003}, and the use of particle filtering
for tracking multiple targets is demonstrated in \cite{MacCormick2000,Hue2001,Vermaak2005}.
But so far, the technique has not been applied to sound source tracking.
Our work demonstrates that it is possible to track multiple sound
sources using particle filters by solving the source-observation assignment
problem.

\section{System Overview}

\label{sec:System-overview}
\begin{figure}[th]
\center{\includegraphics[width=0.7\columnwidth,keepaspectratio]{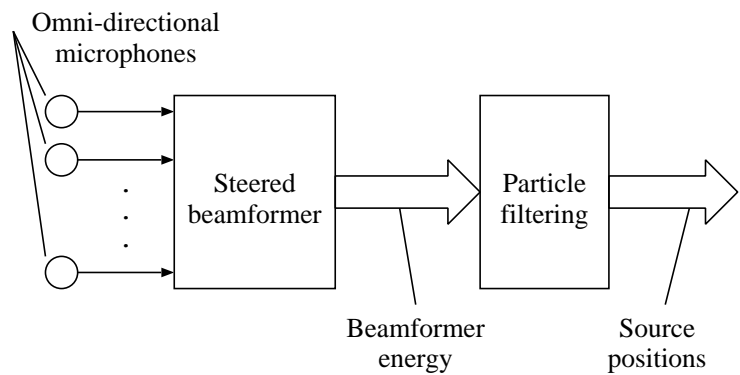}}

\caption{Overview of the localization system\label{cap:Overview-of-the-system}}
\end{figure}

The proposed localization and tracking
system, as shown in Figure \ref{cap:Overview-of-the-system}, is composed
of three parts:

\begin{itemize}
\item A microphone array;
\item A memoryless localization algorithm based on a steered beamformer;
\item A particle filtering tracker.
\end{itemize}
The array is composed of up to eight omni-directional microphones
mounted on the robot. Since the system is designed to be installed
on any robot, there is no strict constraint on the position of the
microphones: only their positions must be known in relation to each
other (measured with $\sim$0.5 cm accuracy). The microphone signals
are used by a beamformer (spatial filter) that is steered in all possible
directions in order to maximize the output energy. The initial localization
performed by the steered beamformer is then used as the input of a
post-processing stage that uses particle filtering to simultaneously
track all sources and prevent false detections. The output of the
localization system can be used to direct the robot attention to the
source. It can also be used as part of a source separation algorithm
to isolate the sound coming from a single source \cite{ValinICRA2004}.

\section{Localization Using a Steered Beamformer}

\label{sec:Localization-by-steered}The basic idea behind the steered
beamformer approach to source localization is to direct a beamformer
in all possible directions and look for maximal output. This can be
done by maximizing the output energy of a simple delay-and-sum beamformer.
The formulation in both time and frequency domain is presented in
Section \ref{sub:Delay-And-Sum-Beamformer}. Section \ref{sub:Spectral-weighting}
describes the frequency-domain weighting performed on the microphone
signals and Section \ref{sub:Direction-Search} shows how the search
is performed. A possible modification for improving the resolution
is described in Section \ref{sub:Direction-Refining}.

\subsection{Delay-And-Sum Beamformer\label{sub:Delay-And-Sum-Beamformer}}

The output of an $M$-microphone delay-and-sum beamformer is defined
as:

\begin{equation}
y(n)=\sum_{m=0}^{M-1}x_{m}\left(n-\tau_{m}\right)\label{eq:delay-and-sum}\end{equation}
where $x_{m}\left(n\right)$ is the signal from the $m^{th}$ microphone
and $\tau_{m}$ is the delay of arrival for that microphone. The output
energy of the beamformer over a frame of length $L$ is thus given
by:\begin{eqnarray}
E & = & \sum_{n=0}^{L-1}\left[y(n)\right]^{2}\nonumber \\
 & = & \sum_{n=0}^{L-1}\left[x_{0}\left(n-\tau_{0}\right)+\ldots+x_{M-1}\left(n-\tau_{M-1}\right)\right]^{2}\label{eq:beamformer-energy}\end{eqnarray}

Assuming that only one sound source is present, we can see that $E$
will be maximal when the delays $\tau_{m}$ are such that the microphone
signals are in phase, and therefore add constructively.

One problem with this technique is that energy peaks are very wide
\cite{Duraiswami2001}, which means that the resolution is poor. Moreover,
in the case where multiple sources are present, it is likely for the
two or more energy peaks to overlap, making them impossible to differentiate.
One way to narrow the peaks is to whiten the microphone signals prior
to computing the energy \cite{Omologo}. Unfortunately, the coarse-fine
search method as proposed in \cite{Duraiswami2001} cannot be used
in that case because the narrow peaks can then be missed during the
coarse search. Therefore, a full fine search is necessary, which requires
increased computing power. It is possible to reduce the amount of
computation by calculating the beamformer energy in the frequency
domain. This also has the advantage of making the whitening of the
signal easier.

To do so, the beamformer output energy in Equation \ref{eq:beamformer-energy}
can be expanded as:

\begin{eqnarray}
E & = & \sum_{m=0}^{M-1}\sum_{n=0}^{L-1}x_{m}^{2}\left(n-\tau_{m}\right)\nonumber \\
 & + & 2\sum_{m_{1}=0}^{M-1}\sum_{m_{2}=0}^{m_{1}-1}\sum_{n=0}^{L-1}x_{m_{1}}\left(n-\tau_{m_{1}}\right)x_{m_{2}}\left(n-\tau_{m_{2}}\right)\label{eq:beamformer-energy-expand}\end{eqnarray}
which in turn can be rewritten in terms of cross-correlations:

\begin{equation}
E=K+2\sum_{m_{1}=0}^{M-1}\sum_{m_{2}=0}^{m_{1}-1}R_{x_{m_{1}},x_{m_{2}}}\left(\tau_{m_{1}}-\tau_{m_{2}}\right)\label{eq:energy-xcorr}\end{equation}
where $K=\sum_{m=0}^{M-1}\sum_{n=0}^{L-1}x_{m}^{2}\left(n-\tau_{m}\right)$
is nearly constant with respect to the $\tau_{m}$ delays and can
thus be ignored when maximizing $E$. The cross-correlation function
can be approximated in the frequency domain as:

\begin{equation}
R_{ij}(\tau)\approx\sum_{k=0}^{L-1}X_{i}(k)X_{j}(k)^{*}e^{\jmath2\pi k\tau/L}\label{eq:TDOA_correlation_freq}\end{equation}
where $X_{i}(k)$ is the discrete Fourier transform of $x_{i}[n]$,
$X_{i}(k)X_{j}(k)^{*}$ is the cross-spectrum of $x_{i}[n]$ and $x_{j}[n]$
and $(\cdot)^{*}$ denotes the complex conjugate. The power spectra
and cross-power spectra are computed on overlapping windows (50\%
overlap) of $L=1024$ samples at 48 kHz. The cross-correlations $R_{ij}(\tau)$
are computed by averaging the cross-power spectra $X_{i}(k)X_{j}(k)^{*}$
over a time period of 4 frames (40 ms). Once the $R_{ij}(\tau)$ are
pre-computed, it is possible to compute $E$ using only $M(M-1)/2$
lookup and accumulation operations, whereas a time-domain computation
would require $2L(M+2)$ operations. For $M=8$ and 2562 directions,
it follows that the complexity of the search itself is reduced from
1.2 Gflops to only 1.7 Mflops. After counting all time-frequency transformations,
the complexity is only 48.4 Mflops, 25 times less than a time domain
search with the same resolution.

\subsection{Spectral Weighting\label{sub:Spectral-weighting}}

In the frequency domain, the whitened cross-correlation is computed
as:\begin{equation}
R_{ij}^{(w)}(\tau)\approx\sum_{k=0}^{L-1}\frac{X_{i}(k)X_{j}(k)^{*}}{\left|X_{i}(k)\right|\left|X_{j}(k)\right|}e^{\jmath2\pi k\tau/L}\label{eq:TDOA_correlation_whitened}\end{equation}

While it produces much sharper cross-correlation peaks, the whitened
cross-correla\-tion has one drawback: each frequency bin of the spectrum
contributes the same amount to the final correlation, even if the
signal at that frequency is dominated by noise. This makes the system
less robust to noise, while making detection of voice (which has a
narrow bandwidth) more difficult. In order to alleviate the problem,
we introduce a weighting function that acts as a mask based on the
signal-to-noise ratio (SNR). For microphone $i$, we define this weighting
function as:\begin{equation}
\zeta_{i}^{n}(k)=\frac{\xi_{i}^{n}(k)}{\xi_{i}^{n}(k)+1}\label{eq:SNR-weighting}\end{equation}
where $\xi_{i}^{n}(k)$ is an estimate of the \emph{a priori} SNR
at the $i^{th}$ microphone, at time frame $n$, for frequency $k$.
It is computed using the decision-directed approach proposed by Ephraim
and Malah \cite{EphraimMalah1984}:\begin{equation}
\xi_{i}^{n}(k)=\frac{(1-\alpha_{d})\left[\zeta_{i}^{n-1}(k)\right]^{2}\left|X_{i}^{n-1}(k)\right|^{2}+\alpha_{d}\left|X_{i}^{n}(k)\right|^{2}}{\sigma_{i}^{2}(k)}\label{eq:decision-directed}\end{equation}
where $\alpha_{d}=0.1$ is the adaptation rate and $\sigma_{i}^{2}(k)$
is the noise estimate for microphone $i$. It is easy to estimate
$\sigma_{i}^{2}(k)$ using the Minima-Controlled Recursive Average
(MCRA) technique \cite{CohenNonStat2001}, which adapts the noise
estimate during periods of low energy. 

It is also possible to make the system more robust to reverberation
by modifying the weighting function in Equation \ref{eq:decision-directed}
to use a new noise estimate $\tilde{\sigma}_{i}^{2}(k)$ that includes
a reverberation term $\lambda_{n,i}^{rev}(k)$ and defined as:\begin{equation}
\tilde{\sigma}_{i}^{2}(k)=\sigma_{i}^{2}(k)+\lambda_{n,i}^{rev}(k)\label{eq:new_noise_estimate}\end{equation}
We use a simple reverberation model with exponential decay defined
as: \begin{equation}
\lambda_{n,i}^{rev}(k)=\gamma\lambda_{n-1,i}^{rev}(k)+(1-\gamma)\delta\left|\zeta_{i}^{n}(k)X_{i}^{n-1}(k)\right|^{2}\label{eq:Reverb_weighting}\end{equation}
where $\gamma$ represents the reverberation decay for the room, $\delta$
is the level of reverberation and $\lambda_{-1,i}^{rev}(k)=0$. In
some sense, Equation \ref{eq:Reverb_weighting} can be seen as modeling
the \emph{precedence effect} \cite{Huang1997Echo,Huang1999Echo} in
order to give less weight to frequency bins where a loud sound was
recently present. The resulting enhanced cross-correlation is defined
as:\begin{equation}
R_{ij}^{(e)}(\tau)=\sum_{k=0}^{L-1}\frac{\zeta_{i}(k)X_{i}(k)\zeta_{j}(k)X_{j}(k)^{*}}{\left|X_{i}(k)\right|\left|X_{j}(k)\right|}e^{\jmath2\pi k\tau/L}\label{eq:TDOA_correlation_weighted}\end{equation}

\subsection{Direction Search on a Spherical Grid\label{sub:Direction-Search}}

\begin{figure}[th]
\center{\includegraphics[width=0.29\columnwidth,height=0.29\columnwidth]{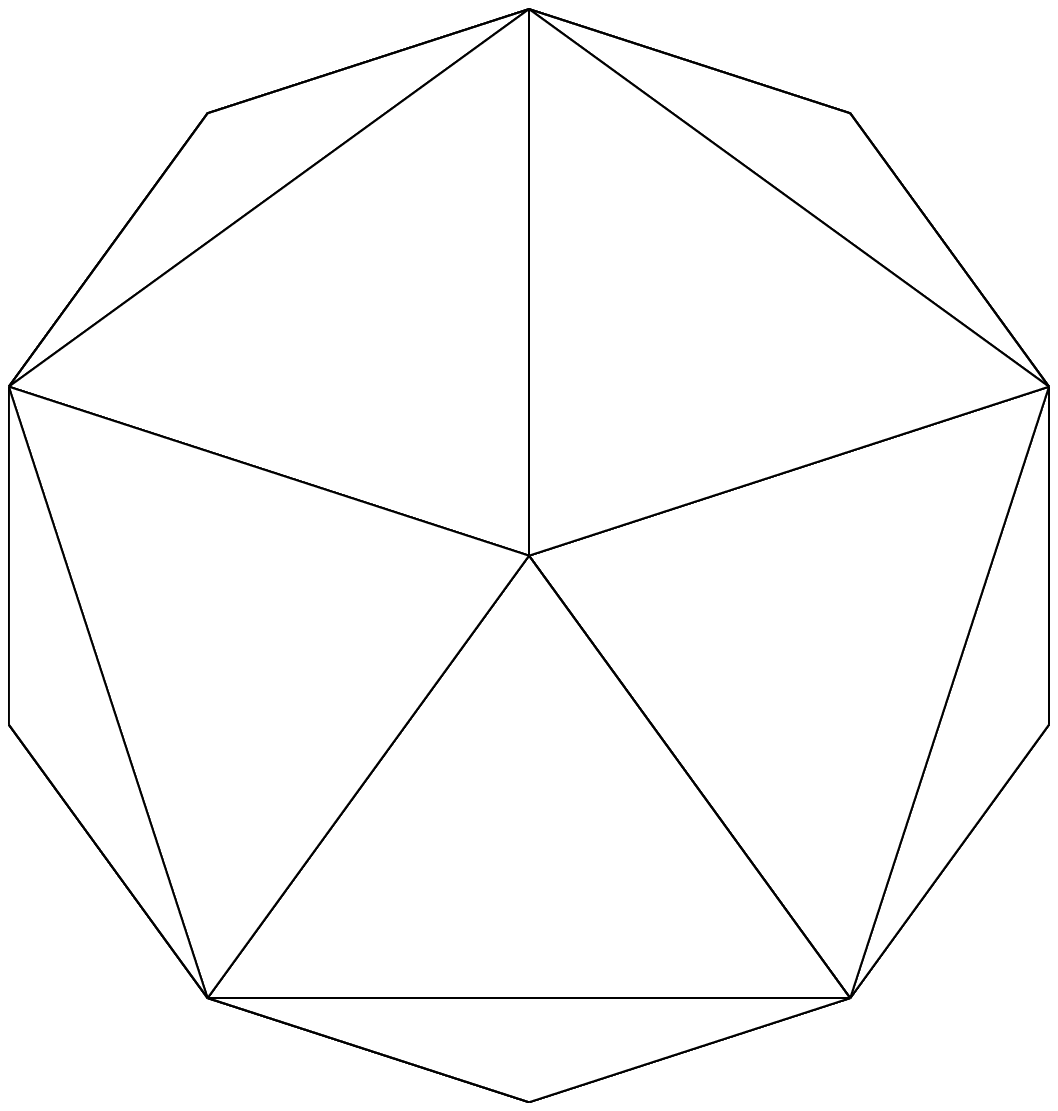}\hspace{3mm}\includegraphics[width=0.29\columnwidth,height=0.29\columnwidth]{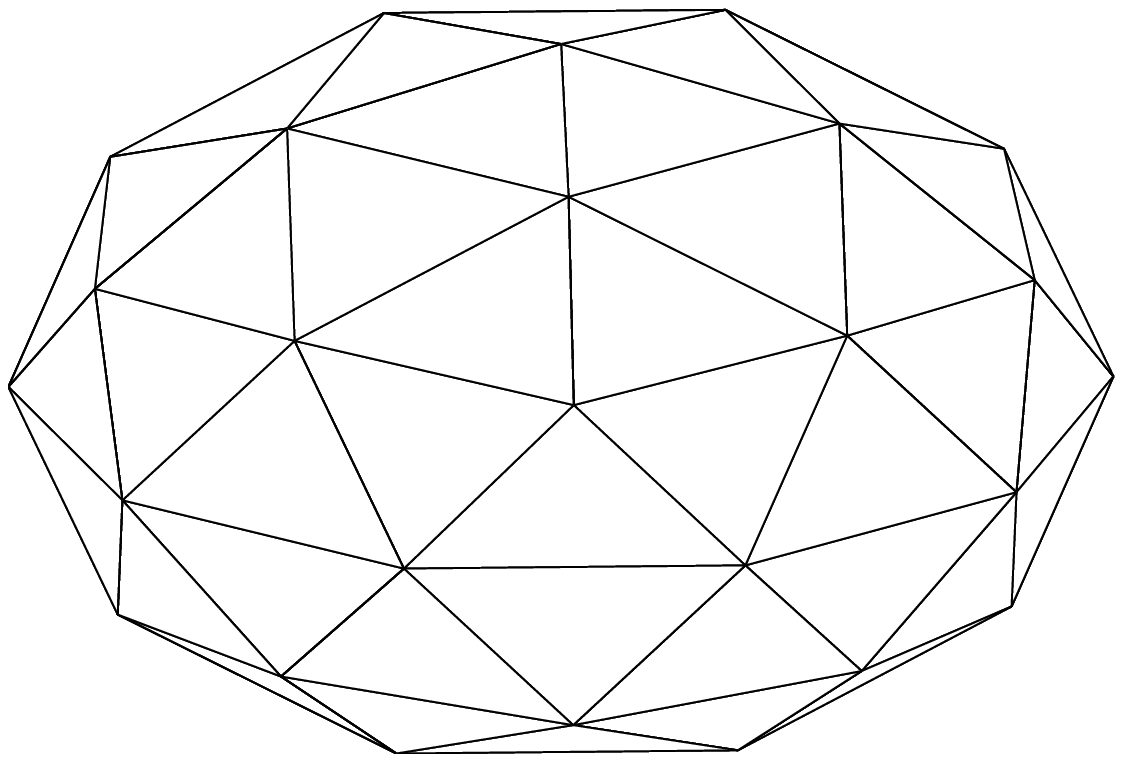}\includegraphics[width=0.32\columnwidth,height=0.32\columnwidth]{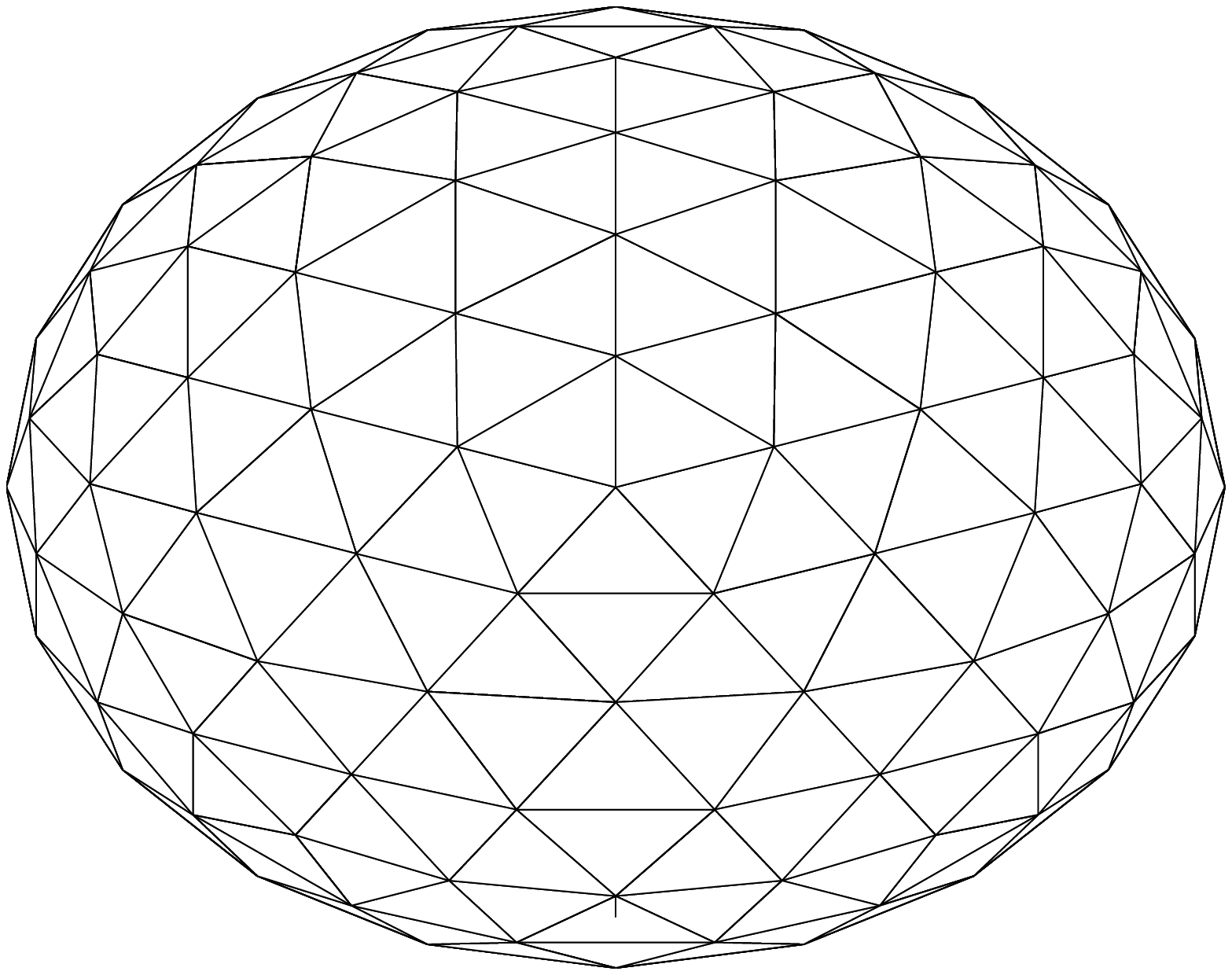}}

\caption{Recursive subdivision (2 levels) of a triangular element\label{cap:Recursive-subdivision}}
\end{figure}

In order to reduce the computation required and to make the system
isotropic, we define a uniform triangular grid for the surface of
a sphere. To create the grid, we start with an initial icosahedral
grid \cite{Giraldo97}. Each triangle in the initial 20-element grid
is recursively subdivided into four smaller triangles, as shown in
Figure \ref{cap:Recursive-subdivision}. The resulting grid is composed
of 5120 triangles and 2562 points. The beamformer energy is then computed
for the hexagonal region associated with each of these points. Each
of the 2562 regions covers a radius of about $2.5^{\circ}$ around
its center, setting the resolution of the search.

\begin{algorithm}[th]
\begin{algorithmic}
\FORALL{grid index $d$}
\STATE $E_d \leftarrow 0$
\FORALL{microphone pair $ij$}
\STATE $\tau \leftarrow lookup(d,ij)$
\STATE $E_d \leftarrow E_d + R^{(e)}_{ij}(\tau)$
\ENDFOR
\ENDFOR
\STATE \textit{direction of source} $\leftarrow \textrm{argmax}_d\ (E_d)$
\end{algorithmic}

\caption{Steered beamformer direction search\label{cap:Steered-beamformer-direction}}
\end{algorithm}

Once the cross-correlations $R_{ij}^{(e)}(\tau)$ are computed, the
search for the best direction on the grid is performed as described
by Algorithm \ref{cap:Steered-beamformer-direction}. The \emph{lookup}
parameter is a pre-computed table of the time delay of arrival (TDOA)
for each microphone pair and each direction on the sphere. Using the
far-field assumption \cite{ValinIROS2003}, the TDOA in samples is
computed as:\begin{equation}
\tau_{ij}=\frac{F_{s}}{c}\left(\vec{\mathrm{\mathbf{p}}}_{i}-\mathrm{\vec{\mathrm{\mathbf{p}}}}_{j}\right)\cdot\vec{\mathbf{u}}\label{eq:TDOA-far-field}\end{equation}
where $\vec{\mathrm{\mathbf{p}}}_{i}$ is the position of microphone
$i$, $\vec{\mathbf{u}}$ is a unit-vector that points in the direction
of the source, $c$ is the speed of sound and $F_{s}$ is the sampling
rate. Equation \ref{eq:TDOA-far-field} assumes that the time delay
is proportional to the distance between the source and microphone.
This is only true when there is no diffraction involved. While this
hypothesis is only verified for an {}``open'' array (all microphones
are in line of sight with the source), in practice we demonstrate
experimentally (see Section \ref{sec:Results}) that the approximation
is good enough for our system to work for a {}``closed'' array (in
which there are obstacles within the array).

For an array of $M$ microphones and an $N$-element grid, the algorithm
requires $M(M-1)N$ table memory accesses and $M(M-1)N/2$ additions.
In the proposed configuration ($N=2562$, $M=8$), the accessed data
can be made to fit entirely in a modern processor's L2 cache.

\begin{algorithm}[th]
\begin{algorithmic}
\FOR{$q=1$ to assumed number of sources}
\STATE $D_q \leftarrow \textrm{Steered beamformer direction search}$
\FORALL{microphone pair $ij$}
\STATE $\tau \leftarrow lookup(D_q,ij)$
\STATE $R^{(e)}_{ij}(\tau) = 0$
\ENDFOR
\ENDFOR
\end{algorithmic}

\caption{Localization of multiple sources\label{cap:Localization-of-multiple-sources}}
\end{algorithm}

Using Algorithm \ref{cap:Steered-beamformer-direction}, our system
is able to find the loudest source present by maximizing the energy
of a steered beamformer. In order to localize other sources that may
be present, the process is repeated by removing the contribution of
the first source to the cross-correlations, leading to Algorithm \ref{cap:Localization-of-multiple-sources}.
Since we do not know how many sources are present, we always look
for four sources, as this is the maximum number of sources our beamformer
is able to locate at once. This situation leads to a high rate of
false detection, even when four or more sources are present. That
problem is handled by the particle filter described in Section \ref{sec:Probabilistic-post-processing}.

\subsection{Direction Refining\label{sub:Direction-Refining}}

When a source is located using Algorithm \ref{cap:Steered-beamformer-direction},
the direction accuracy is limited by the size of the grid used. It
is however possible, as an optional step, to further refine the source
location estimate. In order to do so, we define a refined grid for
the surrounding of the point where a source was found. To take into
account the near-field effects, the grid is refined in three dimensions:
horizontally, vertically and over distance. Using five points in each
direction, we obtain a 125-point local grid with a maximum resolution
error of around $1^{\circ}$. For the near-field case, Equation \ref{eq:TDOA-far-field}
no longer holds, so it is necessary to compute the time differences
as:\begin{equation}
\tau_{ij}=\frac{F_{s}}{c}\left(\left\Vert d\vec{\mathbf{u}}-\vec{\mathbf{p}}_{j}\right\Vert -\left\Vert d\vec{\mathbf{u}}-\vec{\mathbf{p}}_{i}\right\Vert \right)\label{eq:TDOA-near-field}\end{equation}
where $d$ is the distance between the source and the center of the
array. Equation \ref{eq:TDOA-near-field} is evaluated for five distances
$d$ (ranging from 50 cm to 5 m) in order to find the direction of
the source with improved accuracy. Unfortunately, it was observed
that the value of $d$ found in the search is too unreliable to provide
a good estimate of the distance between the source and the array.
The incorporation of the distance nonetheless provides improved accuracy
for the near field case.

\section{Particle-Based Tracking\label{sec:Probabilistic-post-processing}}

The steered beamformer detailed in Section \ref{sec:Localization-by-steered}
provides only instantaneous, noisy information about sources being
possibly present and provides no information about the behavior of
the source in time (tracking). For that reason, it is desirable to
use a probabilistic temporal integration to track the different sound
sources based on all measurements available up to the current time.
It has been shown \cite{Ward2002b,Ward2003,Asoh2004} that particle
filters are an effective way of tracking sound sources. Using this
approach, all hypotheses about the location of each source are represented
as a set of particles to which different weights are assigned.

At time $t$, we consider the case of sources $j=0,1,\ldots,M-1$,
each modeled using $N$ particles of directions $\mathbf{x}_{j,i}^{(t)}$
and weights $w_{j,i}^{(t)}$. The state vector for the particles is
composed of six dimensions, three for position and three for its derivative:
\begin{equation}
\mathbf{s}_{j,i}^{(t)}=\left[\begin{array}{c}
\mathbf{x}_{j,i}^{(t)}\\
\mathbf{\dot{x}}_{j,i}^{(t)}\end{array}\right]\label{eq:particle_state}\end{equation}

\begin{algorithm}[!t]
\begin{enumerate}
\item Predict the state $\mathbf{s}_{j}^{(t)}$ from $\mathbf{s}_{j}^{(t-1)}$
for each source $j$
\item Compute instantaneous direction probabilities associated with the
steered beamformer response
\item Compute probabilities $P_{q,j}^{(t)}$ associating beamformer peaks
to sources being tracked
\item Compute updated particle weights $w_{j,i}^{(t)}$
\item Add or remove sources if necessary
\item Compute source localization estimate $\bar{\mathbf{x}}_{j}^{(t)}$
for each source
\item Resample particles for each source if necessary and go back to step
1.
\end{enumerate}

\caption{Particle-based tracking algorithm. Steps 1 to 7 correspond to Subsections
\ref{sub:Prediction} to \ref{sub:Resampling}.\label{cap:Particle-based-tracking-algorithm}}
\end{algorithm}

Since the particle position is constrained to lie on a unit sphere
and the speed is tangent to the sphere, there are only four degrees
of freedom. The sampling importance resampling (SIR) particle filtering
algorithm is outlined in Figure \ref{cap:Particle-based-tracking-algorithm}
and generalizes sound source tracking to an arbitrary and non-constant
number of sources. The probability density function (pdf) for the
location of each source is approximated by a set of particles that
are given different weights. The weights are updated by taking into
account observations obtained from the steered beamformer and by computing
the assignment between these observations and the sources being tracked.
From there, the estimated location of the source is the weighted mean
of the particle positions.

\subsection{Prediction\label{sub:Prediction}}

As a predictor, we use the excitation-damping model as proposed in
\cite{Ward2003} because it has been observed to work well in practice
and can easily model different source dynamics only two parameters.
The model is defined as:

\begin{eqnarray}
\mathbf{\dot{x}}_{j,i}^{(t)} & = & a\mathbf{\dot{x}}_{j,i}^{(t-1)}+bF_{\mathbf{x}}\label{eq:predict_speed}\\
\mathbf{x}_{j,i}^{(t)} & = & \mathbf{x}_{j,i}^{(t-1)}+\Delta T\mathbf{\dot{x}}_{j,i}^{(t)}\label{eq:predict_pos}\end{eqnarray}
where $a=e^{-\alpha\Delta T}$ controls the damping term, $b=\beta\sqrt{1-a^{2}}$
controls the excitation term, $F_{\mathbf{x}}$ is a normally distributed
random variable of unit variance and $\Delta T$ is the time interval
between updates. We consider three possible states:

\begin{itemize}
\item Stationary source ($\alpha=2$, $\beta=0.04$);
\item Constant velocity source ($\alpha=0.05$, $\beta=0.2$);
\item Accelerated source ($\alpha=0.5$, $\beta=0.2$).
\end{itemize}
A normalization step ensures that $\mathbf{x}_{i}^{(t)}$ still lies
on the unit sphere ($\left\Vert \mathbf{x}_{j,i}^{(t)}\right\Vert =1$)
after applying Equations \ref{eq:predict_speed} and \ref{eq:predict_pos}.

\subsection{Instantaneous Direction Probabilities from Beamformer Response}

\begin{figure}[th]
\center{\includegraphics[width=3.5in,keepaspectratio]{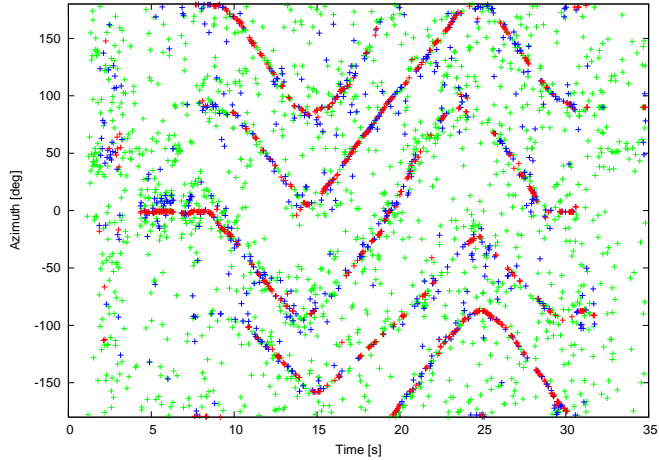}}

\caption{Beamformer output probabilities $P_{q}$ for azimuth as a function
of time. Observations with $P_{q}>0.5$ shown in red, $0.2<P_{q}<0.5$
in blue, $P_{q}<0.2$ in green.\label{cap:Beamformer-output-probabilities}}
\end{figure}

The steered beamformer described in Section \ref{sec:Localization-by-steered}
produces an observation $O^{(t)}$ for each time $t$. The observation
$O^{(t)}=\left[O_{0}^{(t)}\ldots O_{Q-1}^{(t)}\right]$ is composed
of $Q$ potential source locations $\mathbf{y}_{q}$ found by Algorithm
\ref{cap:Localization-of-multiple-sources}. We also denote $\mathbf{O}^{(t)}$,
the set of all observations $O^{(t)}$ up to time $t$. We introduce
the probability $P_{q}$ that the potential source $q$ is a true
source (not a false detection). The value of $P_{q}$ can be interpreted
as our confidence in the steered beamformer output. We know that the
higher the beamformer energy, the more likely a potential source is
to be true. For $q>0$, false alarms are very frequent and independent
of energy. With this in mind, we define $P_{q}$ empirically as:\begin{equation}
P_{q}=\left\{ \begin{array}{ll}
\nu^{2}/2, & q=0,\nu\leq1\\
1-\nu^{-2}/2,\qquad & q=0,\nu>1\\
0.3, & q=1\\
0.16, & q=2\\
0.03, & q=3\end{array}\right.\label{eq:potential-source-reliability}\end{equation}

with $\nu=E_{0}/E_{T}$, where $E_{0}$ is the beamformer output energy
for the first source found and $E_{T}$ is a threshold that depends
on the number of microphones, the frame size and the analysis window
used (we use $E_{T}=150$). Figure \ref{cap:Beamformer-output-probabilities}
shows an example of $P_{q}$ values for potential sources found by
the steered beamformer with four people speaking continuously while
moving around the microphone array in a moderately reverberant room.
Only the azimuth part of $\mathbf{y}_{q}$ is shown as a function
of time.

At time $t$, the probability density of observing $O_{q}^{(t)}$
for a source located at particle position $\mathbf{x}_{j,i}^{(t)}$
is given by:\begin{equation}
p\left(\left.O_{q}^{(t)}\right|\mathbf{x}_{j,i}^{(t)}\right)=\mathcal{N}\left(\mathbf{y}_{q};\mathbf{x}_{j,i};\sigma^{2}\right)\label{eq:normal-pdf}\end{equation}
where $\mathcal{N}\left(\mathbf{y}_{q};\mathbf{x}_{j,i};\sigma^{2}\right)$
is a normal distribution centered at $\mathbf{x}_{j,i}$ with variance
$\sigma^{2}$ evaluated at $\mathbf{y}_{q}$, and models the localization
accuracy of the steered beamformer. We use $\sigma=0.05$, which corresponds
to an RMS error of 3 degrees for the location found by the steered
beamformer. This error takes into account the resolution error (1
degree) as well as other sources of errors, such as noise, reverberation,
diffraction, imperfect microphones and errors in microphone placement.

\subsection{Probabilities for Multiple Sources}

Before we can derive the update rule for the particle weights $w_{j,i}^{(t)}$,
we must first introduce the concept of source-observation assignment.
For each potential source $q$ detected by the steered beamformer,
there are three possibilities:

\begin{itemize}
\item It is a false detection ($H_{0}$).
\item It corresponds to one of the sources currently tracked ($H_{1}$).
\item It corresponds to a new source that is not yet being tracked ($H_{2}$).
\end{itemize}
In the case of $H_{1}$, we need to determine which tracked source
$j$ corresponds to potential source $q$. First, we assume that a
potential source may correspond to at most one tracked source and
that a tracked source can correspond to at most one potential source. 

\begin{figure}[h]
\center{\includegraphics[width=5cm,keepaspectratio]{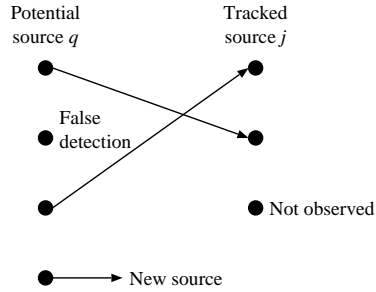}}

\caption{Assignment example where two of the tracked sources are observed,
with one new source and one false detection. The assignment can be
described as $f(\{0,1,2,3\})=\{1,-2,0,-1\}$.\label{cap:Mapping-example}}
\end{figure}

Let $f:\{0,1,\ldots,Q-1\}\longrightarrow\{-2,-1,0,1,\ldots,M-1\}$
be a function assigning observation $q$ to the source $j$ (values
-2 is used for false detection and -1 is used for a new source). Figure
\ref{cap:Mapping-example} illustrates a hypothetical case with four
potential sources detected by the steered beamformer and their assignment
to the tracked sources. Knowing $P\left(f\left|O^{(t)}\right.\right)$
(the probability that $f$ is the correct assignment given observation
$O^{(t)}$) for all possible $f$, we can derive $P_{q,j}$, the probability
that the tracked source $j$ corresponds to the potential source $q$
as:\begin{eqnarray}
P_{q,j}^{(t)} & = & \sum_{f}\delta_{j,f(q)}P\left(f\left|O^{(t)}\right.\right)\label{eq:Pqj}\\
P_{q}^{(t)}\left(H_{0}\right) & = & \sum_{f}\delta_{-2,f(q)}P\left(f\left|O^{(t)}\right.\right)\label{eq:PH0}\\
P_{q}^{(t)}\left(H_{2}\right) & = & \sum_{f}\delta_{-1,f(q)}P\left(f\left|O^{(t)}\right.\right)\label{eq:PH2}\end{eqnarray}
where $\delta_{i,j}$ is the Kronecker delta. Equation \ref{eq:Pqj}
is in fact the sum of the probabilities of all $f$ that assign potential
source $q$ to tracked source $j$ and similarly for Equations \ref{eq:PH0}
and \ref{eq:PH2}.

Omitting $t$ for clarity, the probability $P(f|O)$ is given by:

\begin{equation}
P(f|O)=\frac{p(O|f)P(f)}{p(O)}\label{eq:mapping-prob}\end{equation}
Knowing that there is only one correct assignment ($\sum_{f}P(f|O)=1$),
we can avoid computing the denominator $p(O)$ by using normalization.
Assuming conditional independence of the observations given the mapping
function, we can decompose $p\left(\left.O\right|f\right)$ into individual
components:\begin{equation}
p\left(\left.O\right|f\right)=\prod_{q}p\left(\left.O_{q}\right|f(q)\right)\label{eq:inverse-mapping-prob-prod}\end{equation}
We assume that the distribution of the false detections ($H_{0}$)
and the new sources ($H_{2}$) are uniform, while the distribution
for tracked sources ($H_{1}$) is the pdf approximated by the particle
distribution convolved with the steered beamformer error pdf:\begin{equation}
p\left(\left.O_{q}\right|f(q)\right)=\left\{ \begin{array}{ll}
1/4\pi,\qquad & f(q)=-2\\
1/4\pi, & f(q)=-1\\
\sum_{i}w_{f(q),i}p\left(\left.O_{q}\right|\mathbf{x}_{j,i}\right), & f(q)\geq0\end{array}\right.\label{eq:indep-inverse-mapping}\end{equation}
The \emph{a priori} probability of $f$ being the correct assignment
is also assumed to come from independent individual components, so
that:\begin{equation}
P(f)=\prod_{q}P\left(f(q)\right)\label{eq:apriori-mapping}\end{equation}
with:\begin{equation}
P\left(f(q)\right)=\left\{ \begin{array}{ll}
\left(1-P_{q}\right)P_{false},\qquad & f(q)=-2\\
P_{q}P_{new} & f(q)=-1\\
P_{q}P\left(Obs_{j}^{(t)}\left|\mathbf{O}^{(t-1)}\right.\right) & f(q)\geq0\end{array}\right.\label{eq:indep-apriori-mapping}\end{equation}
where $P_{new}$ is the \emph{a priori} probability that a new source
appears and $P_{false}$ is the \emph{a priori} probability of false
detection. The probability $P\left(Obs_{j}^{(t)}\left|\mathbf{O}^{(t-1)}\right.\right)$
that source $j$ is observable (i.e., that it exists and is active)
at time $t$ is given by:

\begin{equation}
P\left(Obs_{j}^{(t)}\left|\mathbf{O}^{(t-1)}\right.\right)=P\left(E_{j}\left|\mathbf{O}^{(t-1)}\right.\right)P\left(\mathrm{A}_{j}^{(t)}\left|\mathbf{O}^{(t-1)}\right.\right)\label{eq:prob-observable}\end{equation}
where $E_{j}$ is the event that source $j$ actually exists and $A_{j}^{(t)}$
is the event that it is active (but not necessarily detected) at time
$t$. By active, we mean that the signal it emits is non-zero (for
example, a speaker who is not making a pause). The probability that
the source exists is given by:\begin{equation}
P\left(E_{j}\left|\mathbf{O}^{(t-1)}\right.\right)=P_{j}^{(t-1)}+\left(1-P_{j}^{(t-1)}\right)\frac{P_{o}P\left(E_{j}\left|\mathbf{O}^{(t-2)}\right.\right)}{1-\left(1-P_{o}\right)P\left(E_{j}\left|\mathbf{O}^{(t-2)}\right.\right)}\label{eq:prob-exist}\end{equation}
where $P_{o}$ is the \emph{a priori} probability that a source is
not observed (i.e., undetected by the steered beamformer) even if
it exists (with $P_{0}=0.2$ in our case) and $P_{j}^{(t)}=\sum_{q}P_{q,j}^{(t)}$
is the probability that source $j$ is observed (assigned to any of
the potential sources).

Assuming a first order Markov process, we can write the following
about the probability of source activity:\begin{eqnarray}
P\left(\mathrm{A}_{j}^{(t)}\left|\mathbf{O}^{(t-1)}\right.\right) & = & P\left(\mathrm{A}_{j}^{(t)}\left|\mathrm{A}_{j}^{(t-1)}\right.\right)P\left(\mathrm{A}_{j}^{(t-1)}\left|\mathbf{O}^{(t-1)}\right.\right)\nonumber \\
 &  & +P\left(\mathrm{A}_{j}^{(t)}\left|\neg\mathrm{A}_{j}^{(t-1)}\right.\right)\left[1-P\left(\mathrm{A}_{j}^{(t-1)}\left|\mathbf{O}^{(t-1)}\right.\right)\right]\label{eq:prob-active-t-1}\end{eqnarray}
with $P\left(\mathrm{A}_{j}^{(t)}\left|\mathrm{A}_{j}^{(t-1)}\right.\right)$
the probability that an active source remains active (set to 0.95),
and $P\left(\mathrm{A}_{j}^{(t)}\left|\neg\mathrm{A}_{j}^{(t-1)}\right.\right)$
the probability that an inactive source becomes active again (set
to 0.05). Assuming that the active and inactive states are equiprobable,
the activity probability is computed using Bayes' rule and usual probability
manipulations:\begin{equation}
P\left(\mathrm{A}_{j}^{(t)}\left|\mathbf{O}^{(t)}\right.\right)=\frac{1}{1+\frac{\left[1-P\left(\mathrm{A}_{j}^{(t)}\left|\mathbf{O}^{(t-1)}\right.\right)\right]\left[1-P\left(\mathrm{A}_{j}^{(t)}\left|O^{(t)}\right.\right)\right]}{P\left(\mathrm{A}_{j}^{(t)}\left|\mathbf{O}^{(t-1)}\right.\right)P\left(\mathrm{A}_{j}^{(t)}\left|O^{(t)}\right.\right)}}\label{eq:prob-active}\end{equation}

\subsection{Weight Update}

At times $t$, the new particle weights for source $j$ are defined
as:

\begin{equation}
w_{j,i}^{(t)}=p\left(\mathbf{x}_{j,i}^{(t)}\left|\mathbf{O}^{(t)}\right.\right)\label{eq:weight_update_def}\end{equation}

Assuming that the observations are conditionally independent given
the source position, and knowing that for a given source $j$, $\sum_{i=1}^{N}w_{j,i}^{(t)}=1$,
we obtain through Bayesian inference:

\begin{eqnarray}
w_{j,i}^{(t)} & = & \frac{p\left(\left.\mathbf{O}^{(t)}\right|\mathbf{x}_{j,i}^{(t)}\right)p\left(\mathbf{x}_{j,i}^{(t)}\right)}{p\left(\mathbf{O}^{(t)}\right)}\nonumber \\
 & = & \frac{p\left(\left.O^{(t)}\right|\mathbf{x}_{j,i}^{(t)}\right)p\left(\left.\mathbf{O}^{(t-1)}\right|\mathbf{x}_{j,i}^{(t)}\right)p\left(\mathbf{x}_{j,i}^{(t)}\right)}{p\left(\mathbf{O}^{(t)}\right)}\nonumber \\
 & = & \frac{p\left(\mathbf{x}_{j,i}\left|O^{(t)}\right.\right)p\left(\mathbf{x}_{j,i}^{(t)}\left|\mathbf{O}^{(t-1)}\right.\right)p\left(O^{(t)}\right)p\left(\mathbf{O}^{(t-1)}\right)}{p\left(\mathbf{O}^{(t)}\right)p\left(\mathbf{x}_{j,i}^{(t)}\right)}\nonumber \\
 & = & \frac{p\left(\mathbf{x}_{j,i}^{(t)}\left|O^{(t)}\right.\right)w_{j,i}^{(t-1)}}{\sum_{i=1}^{N}p\left(\mathbf{x}_{j,i}^{(t)}\left|O^{(t)}\right.\right)w_{j,i}^{(t-1)}}\label{eq:weight-update}\end{eqnarray}

Let $I_{j}^{(t)}$ denote the event that source $j$ is observed at
time $t$ and knowing that $P\left(I_{j}^{(t)}\right)=P_{j}^{(t)}=\sum_{q}P_{q,j}^{(t)}$,
we have:

\begin{equation}
p\left(\mathbf{x}_{j,i}^{(t)}\left|O^{(t)}\right.\right)=\left(1-P_{j}^{(t)}\right)p\left(\mathbf{x}_{j,i}^{(t)}\left|O^{(t)},\neg I_{j}^{(t)}\right.\right)+P_{j}^{(t)}p\left(\mathbf{x}_{j,i}^{(t)}\left|O^{(t)},I_{j}^{(t)}\right.\right)\label{eq:instant-weight}\end{equation}
In the case where no observation matches the source, all particles
have the same probability, so we obtain:

\begin{equation}
p\left(\mathbf{x}_{j,i}^{(t)}\left|O^{(t)}\right.\right)=\left(1-P_{j}^{(t)}\right)\frac{1}{N}+P_{j}\frac{\sum_{q=1}^{Q}P_{q,j}^{(t)}p\left(\left.O_{q}^{(t)}\right|\mathbf{x}_{j,i}^{(t)}\right)}{\sum_{i=1}^{N}\sum_{q=1}^{Q}P_{q,j}^{(t)}p\left(\left.O_{q}^{(t)}\right|\mathbf{x}_{j,i}^{(t)}\right)}\label{eq:instant-weight2}\end{equation}
where the denominator on the right side of Equation \ref{eq:instant-weight2}
provides normalization for the $I_{j}^{(t)}$ case, so that $\sum_{i=1}^{N}p\left(\mathbf{x}_{j,i}^{(t)}\left|O^{(t)},I_{j}^{(t)}\right.\right)=1$.

\subsection{Adding or Removing Sources}

In a real environment, sources may appear or disappear at any moment.
If, at any time, $P_{q}(H_{2})$ is higher than a threshold equal
to $0.3$, we consider that a new source is present. In that case,
a set of particles is created for source $q$. Even when a new source
is created, it is only assumed to exist if its probability of existence
$P\left(E_{j}\left|\mathbf{O}^{(t)}\right.\right)$ reaches a certain
threshold, which we set to 0.98. At this point, the probability of
existence is set up 1 and ceases to be updated.

In the same way, we set a time limit on sources. If the source has
not been observed ($P_{j}^{(t)}<T_{obs}$) for a certain amount of
time, we consider that it no longer exists. In that case, the corresponding
particle filter is no longer updated nor considered in future calculations.

\subsection{Parameter Estimation}

The estimated position of each source is the mean of the pdf and can
be obtained as a weighted average of its particles position:

\begin{equation}
\bar{\mathbf{x}}_{j}^{(t)}=\sum_{i=1}^{N}w_{j,i}^{(t)}\mathbf{x}_{j,i}^{(t)}\label{eq:pos_estimation_mean}\end{equation}

\begin{figure}[th]
\center{\includegraphics[width=2.5in]{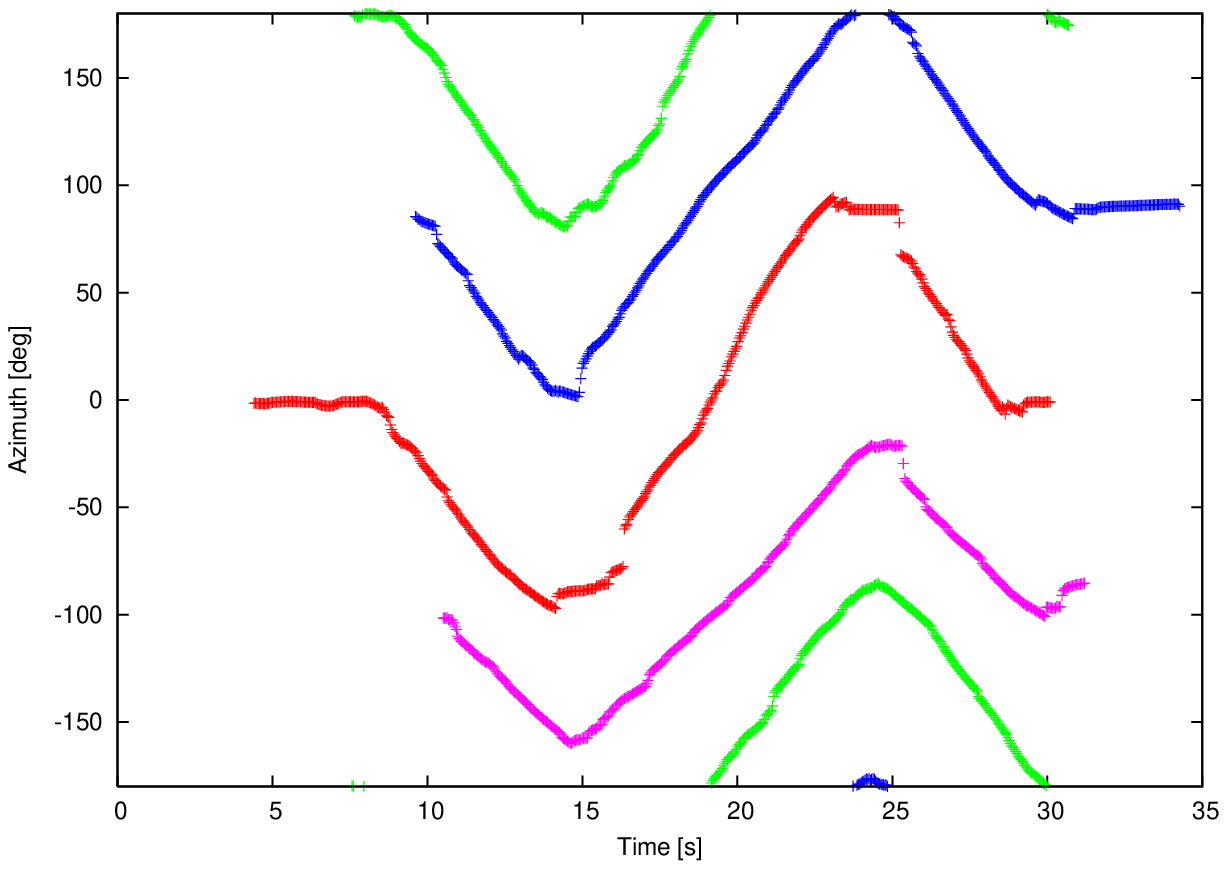}\includegraphics[width=2.5in]{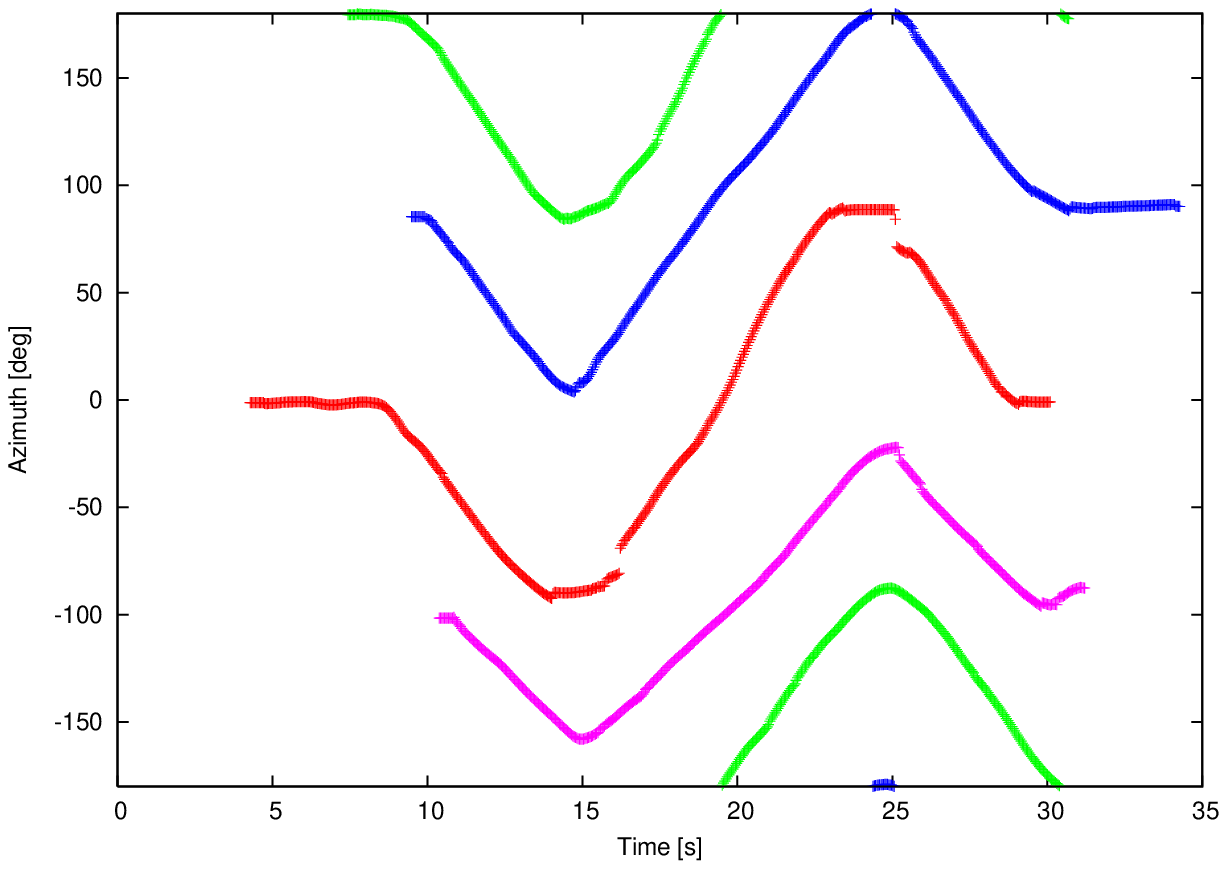}}

\caption{Tracking of four moving sources, showing azimuth as a function of
time. Left: no delay, right: delayed estimation (500 ms).\label{cap:Tracking-delay}}
\end{figure}

It is however possible to obtain better accuracy simply by adding
a delay to the algorithm. This can be achieved by augmenting the state
vector by past position values. At time $t$, the position at time
$t-T$ is thus expressed as:\begin{equation}
\bar{\mathbf{x}}_{j}^{(t-T)}=\sum_{i=1}^{N}w_{j,i}^{(t)}\mathbf{x}_{j,i}^{(t-T)}\label{eq:pos_estimation_delayed}\end{equation}
Using the same example as in Figure \ref{cap:Beamformer-output-probabilities}
we show in Figure \ref{cap:Tracking-delay} how the particle filter
is able to remove the noise and produce smooth trajectories. The added
delay produces an even smoother result.

\subsection{Resampling\label{sub:Resampling}}

Resampling is performed only when $N_{eff}\approx\left(\sum_{i=1}^{N}w_{j,i}^{2}\right)^{-1}<N_{min}$
\cite{Doucet2000} with $N_{min}=0.7N$. That criterion ensures that
resampling only occurs when new data is available for a certain source.
Otherwise, this would cause unnecessary reduction in particle diversity,
due to some particles randomly disappearing.

\section{Results\label{sec:Results}}

\begin{figure}[th]
\center{\includegraphics[height=9cm,keepaspectratio]{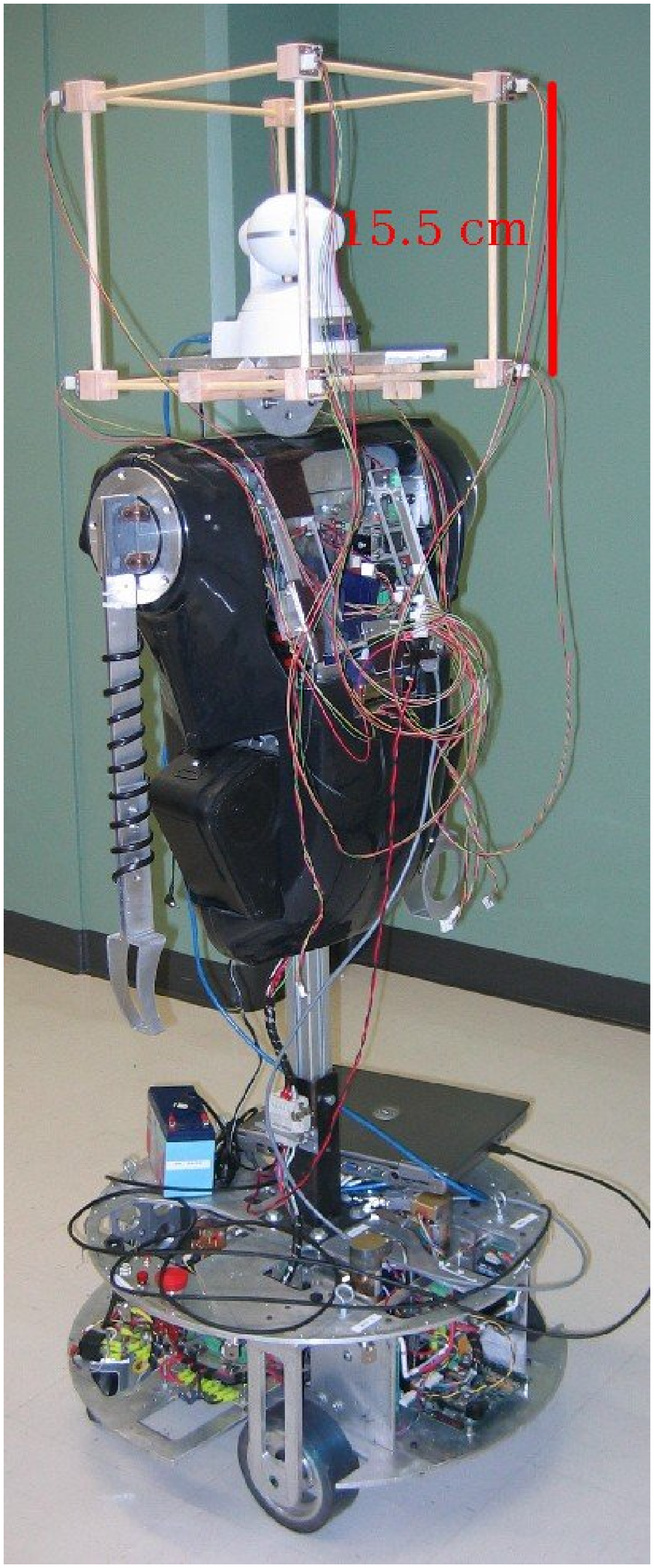}$\;$\includegraphics[height=9cm,keepaspectratio]{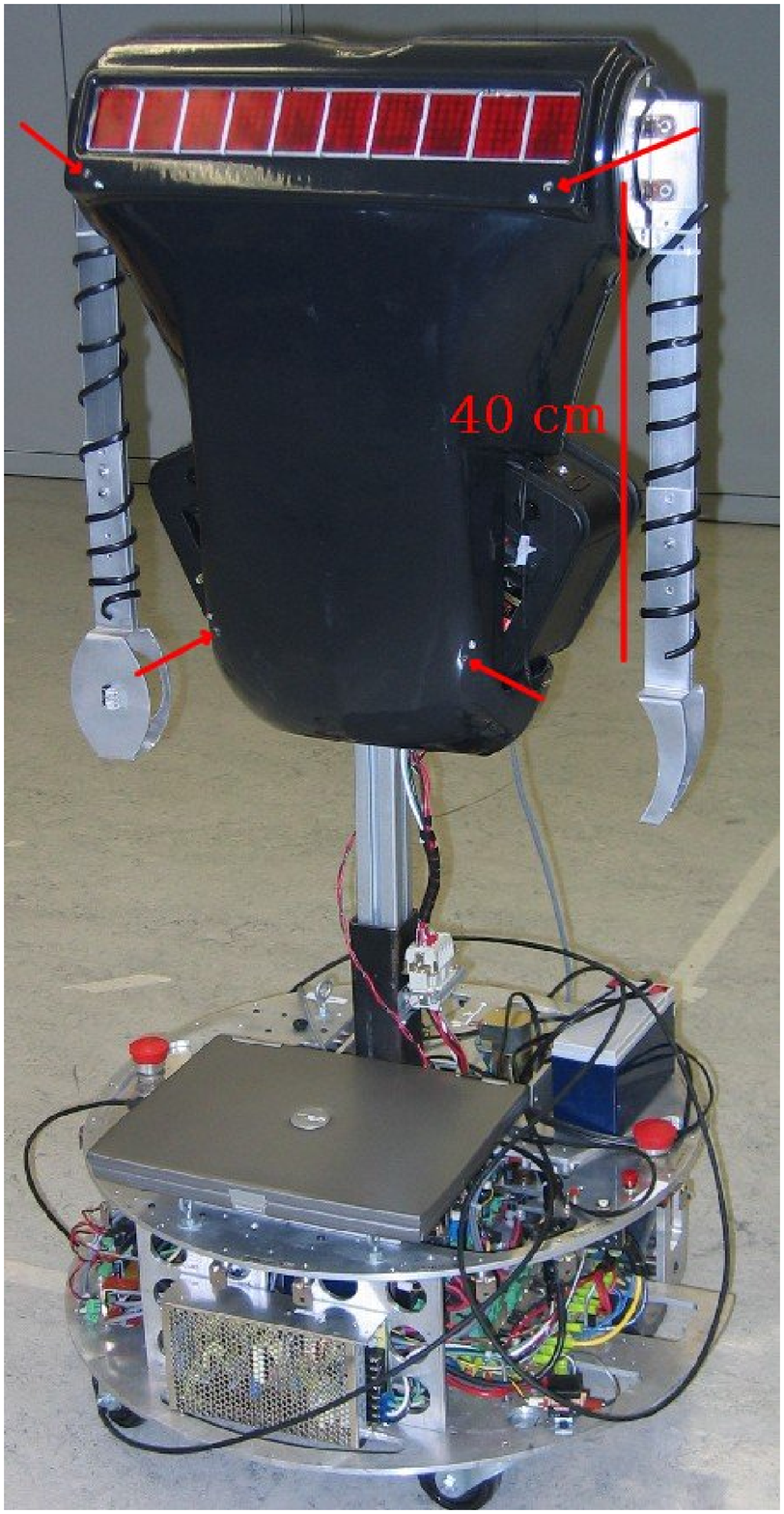}}

\caption{Spartacus robot in configuration C1 (left) and C2 (right)\label{cap:Spartacus-robot}.}
\end{figure}

The proposed localization system is tested using an array of omni-directional
microphones, each composed of an electret cartridge mounted on a simple
pre-amplifier. The array is composed of eight microphones, as it is
the maximum number of analog input channels on commercially available
soundcards. Two array configurations are used for the evaluation of
the system. The first configuration (C1) is an open array and consists
of inexpensive ($\sim$\$1 each) microphones arranged on the summits
of a 16 cm cube mounted on top of the \emph{Spartacus} robot (shown
left in Figure \ref{cap:Spartacus-robot}). The second configuration
(C2) is a closed array and uses smaller, middle-range ($\sim$\$20
each) microphones, placed through holes at different locations on
the body of the robot (shown right in Figure \ref{cap:Spartacus-robot}).
For both arrays, all channels are sampled simultaneously using an
RME Hammerfall Multiface DSP connected to a laptop through a CardBus
interface. Running the localization system in real-time currently
requires 30\% of a 1.6~GHz Pentium-M CPU. Due to the low complexity
of the particle filtering algorithm, we are able to use 1000 particles
per source without noticeable increase in complexity. This also means
that the CPU time does not increase significantly with the number
of sources present. 

Experiments are performed in two different environments. The first
environment (E1) is a medium-size room (10 m $\times$ 11 m, 2.5 m
ceiling) with a reverberation time (-60 dB) of 350 ms. The second
environment (E2) is a hall (16 m $\times$ 17 m, 3.1 m ceiling, connected
to other rooms) with 1.0 s reverberation time. For all tasks, configurations
and environments, all parameters have the same value, except for the
reverberation decay $\gamma$, which is set to 0.65 in the E1 environment
and 0.85 in the E2 environment.

\subsection{Characterization}

The system is characterized in environment E1 in terms of detection
reliability and accuracy. Detection reliability is defined as the
capacity to detect and localize sounds to within 10 degrees, while
accuracy is defined as the localization error for sources that are
detected. We use three different types of sound: a hand clap, the
test sentence {}``Spartacus, come here'', and a burst of white noise
lasting 100 ms. The sounds are played from a speaker placed at different
locations around the robot and at three different heights: 0.1 m,
1 m, 1.4 m.

\subsubsection{Detection Reliability}

Detection reliability is tested at distances (measured from the center
of the array) ranging from $1\:\mathrm{m}$ (a normal distance for
close interaction) to $7\:\mathrm{m}$ (limitation of the room). Three
indicators are computed: correct localization (within 10 degrees),
reflections (incorrect elevation due to roof of ceiling), and other
errors. For all indicators, we compute the number of occurrences divided
by the number of sounds played. This test includes 1440 sounds at
a 22.5$^{\circ}$ interval for 1 m and 3 m and 360 sounds at a 90$^{\circ}$
interval for 5 m and 7 m. Because of the limited size of the room
used for the experiment, the tests for 5 m and 7 m had to use fixed
positions for the robot and the source, leading to less variability
in the conditions. This can explain differences between these results
and those obtained for shorted distances, especially for reflections.

Results are shown in Table \ref{cap:Detection-reliability-C1C2} for
both C1 and C2 configurations. In configuration C1, results show near-perfect
reliability even at seven meter distance. For C2, we noted that the
reliability depends on the sound type, so detailed results for different
sounds are provided in Table \ref{cap:Detection-clap-speech-noise},
showing that only hand clap sounds cannot be reliably detected passed
one meter. We expect that a human would have achieved a score of 100\%
for this reliability test.

Like most localization algorithms, our system is unable to detect
pure tones. This behavior is explained by the fact that sinusoids
occupy only a very small region of the spectrum and thus have a very
small contribution to the cross-correlations with the proposed weighting.
It must be noted that tones tend to be more difficult to localize
even for the human auditory system. 

\begin{table}[th]

\caption{Detection reliability for C1 and C2 configurations\label{cap:Detection-reliability-C1C2}}

\begin{tabular}{|c||c|c||c|c||c|c|}
\hline 
Distance&
\multicolumn{2}{c||}{Correct (\%)}&
\multicolumn{2}{c||}{Reflection (\%)}&
\multicolumn{2}{c|}{Other error (\%)}\tabularnewline
\cline{1-1} \cline{4-5} \cline{6-7} 
\hline 
&
C1&
C2&
C1&
C2&
C1&
C2\tabularnewline
\hline
\hline 
1 m&
100&
94.2&
0.0&
7.3&
0.0&
1.3\tabularnewline
\hline 
 3 m&
99.4&
80.6&
0.0&
21.0&
0.3&
0.1\tabularnewline
\hline 
5 m&
98.3&
89.4&
0.0&
0.0&
0.0&
1.1\tabularnewline
\hline 
7 m&
100&
85.0&
0.6&
1.1&
0.6&
1.1\tabularnewline
\hline
\end{tabular}
\end{table}

\begin{table}[th]

\caption{Correct localization rate as a function of sound type and distance
for C2 configuration\label{cap:Detection-clap-speech-noise}}

\begin{tabular}{|c|c|c|c|}
\hline 
Distance&
Hand clap (\%)&
Speech (\%)&
Noise burst (\%)\tabularnewline
\hline
\hline 
1 m&
88.3&
98.3&
95.8\tabularnewline
\hline 
3 m&
50.8&
97.9&
92.9\tabularnewline
\hline 
5 m&
71.7&
98.3&
98.3\tabularnewline
\hline 
7 m&
61.7&
95.0&
98.3\tabularnewline
\hline
\end{tabular}
\end{table}

\subsubsection{Localization Accuracy}

In order to measure the accuracy of the localization system, we use
the same setup as for measuring reliability, with the exception that
only distances of $1\:\mathrm{m}$ and $3\:\mathrm{m}$ are tested
(1440 sounds at a 22.5$^{\circ}$ interval) due to limited space available
in the testing environment. Neither distance nor sound type has significant
impact on accuracy. The root mean square accuracy results are shown
in Table \ref{cap:Localization-accuracy} for configurations C1 and
C2. Both azimuth and elevation are shown separately. According to
\cite{Hartmann1983,Rakerd2004}, human sound localization accuracy
ranges between two and four degrees in similar conditions. The localization
accuracy of our system is thus equivalent or better than human localization
accuracy.

\begin{table}[th]

\caption{Localization accuracy (root mean square error)\label{cap:Localization-accuracy}}

\begin{tabular}{|c|c|c|}
\hline 
Localization error&
C1 (deg)&
C2 (deg)\tabularnewline
\hline
\hline 
Azimuth&
1.10&
1.44\tabularnewline
\hline 
Elevation&
0.89&
1.41\tabularnewline
\hline
\end{tabular}
\end{table}

\subsection{Source Tracking}

\begin{figure}[th]
\center{\includegraphics[width=1.5in,keepaspectratio]{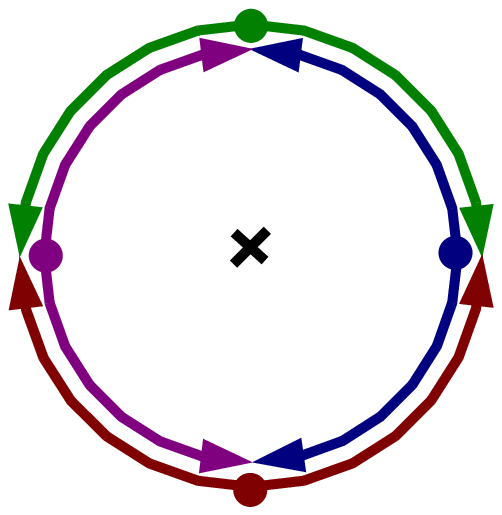}\includegraphics[width=1.5in,keepaspectratio]{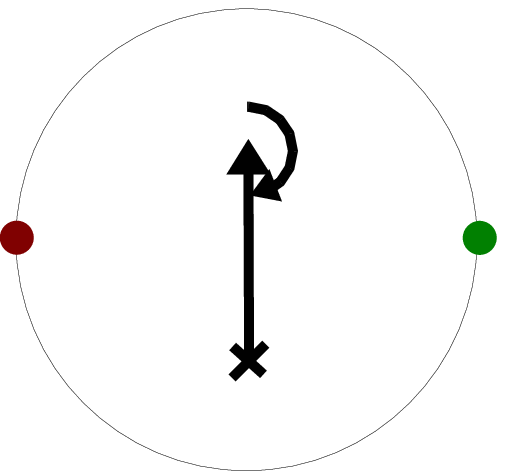}\includegraphics[width=1.5in,keepaspectratio]{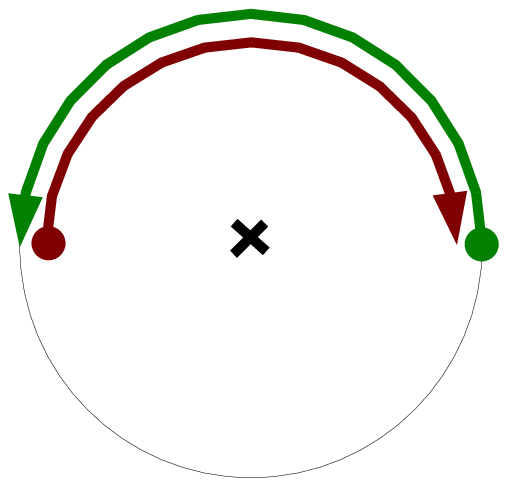}}

\caption{Source trajectories (robot represented as an X). Left: moving sources.
Center: moving robot. Right: sources with intersecting trajectories.\label{cap:Source-trajectories}}
\end{figure}

We measure the tracking capabilities of the system for multiple sound
sources. These are performed using the C2 configuration in both E1
and E2 environments. In all cases, the distance between the robot
and the sources is approximately two meters. The azimuth is shown
as a function of time for each source. The elevation is not shown
as it is almost the same for all sources during these tests. The trajectories
for the three experiments are shown in Figure \ref{cap:Source-trajectories}.

\subsubsection{Moving Sources\label{sub:Moving-Sources}}

In a first experiments, four people were told to talk continuously
(reading a text with normal pauses between words) to the robot while
moving, as shown on the left of Figure \ref{cap:Source-trajectories}.
Each person walked 90 degrees towards the left of the robot before
walking 180 degrees towards the right. 

\begin{figure}[th]
\includegraphics[width=2.5in]{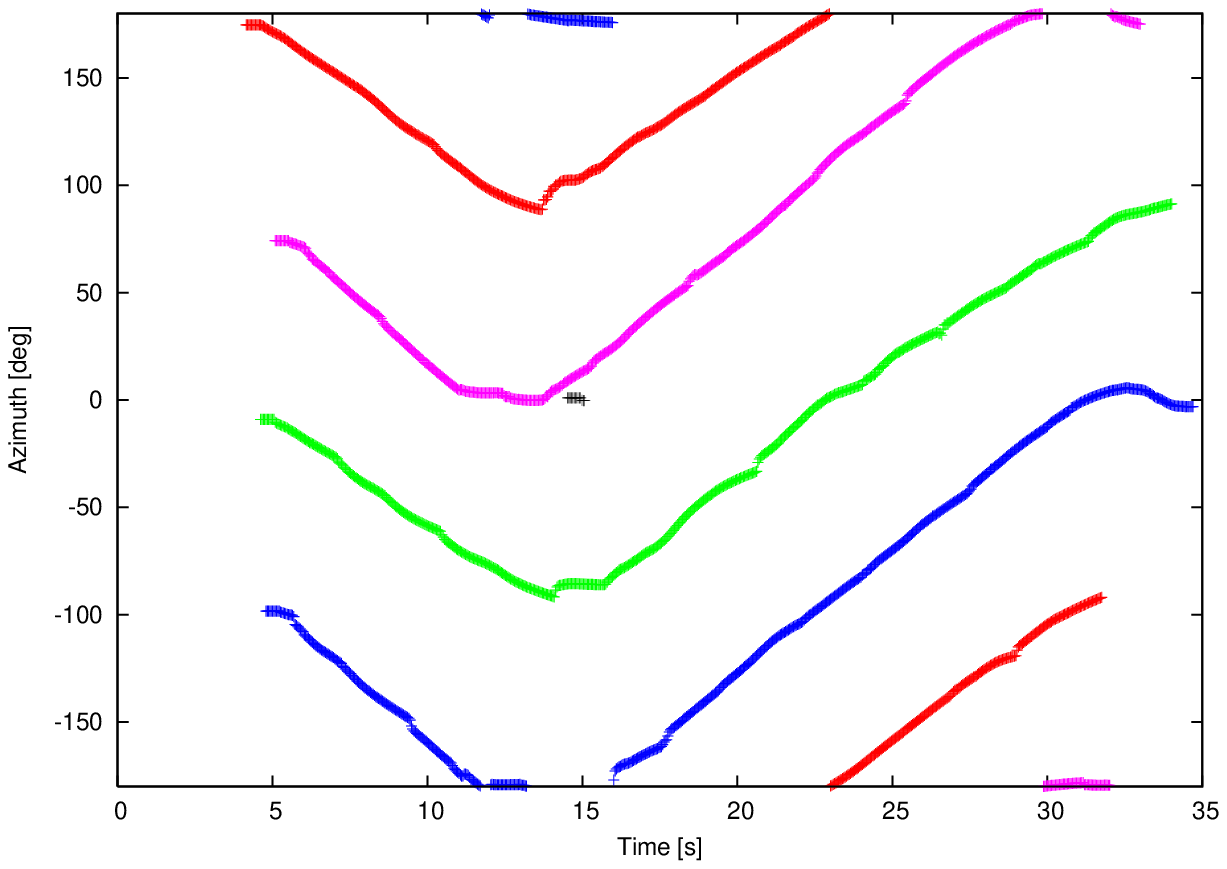}\includegraphics[width=2.5in]{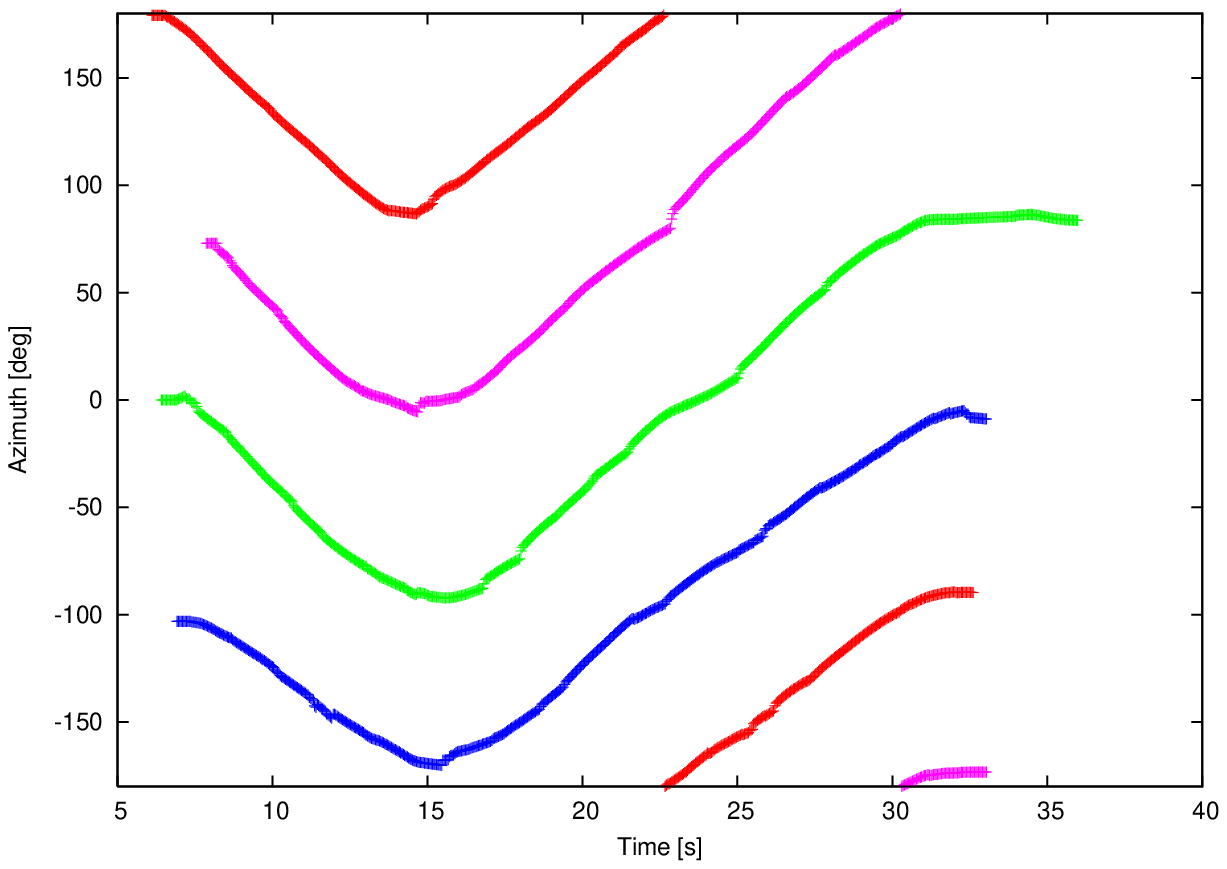}

\caption{Four speakers moving around a stationary robot. Left: E1, right:
E2. False detection shown in black. \label{cap:Four-speakers-moving}}
\end{figure}

Results are presented in Figure \ref{cap:Four-speakers-moving} for
delayed estimation (500 ms). In both environments, the source estimated
trajectories are consistent with the trajectories of the four speakers
and only one false detection was present (in E1, at $t=15\:\mathrm{s}$)
for a short period of time.

\subsubsection{Moving Robot}

\begin{figure}[th]
\includegraphics[width=2.5in]{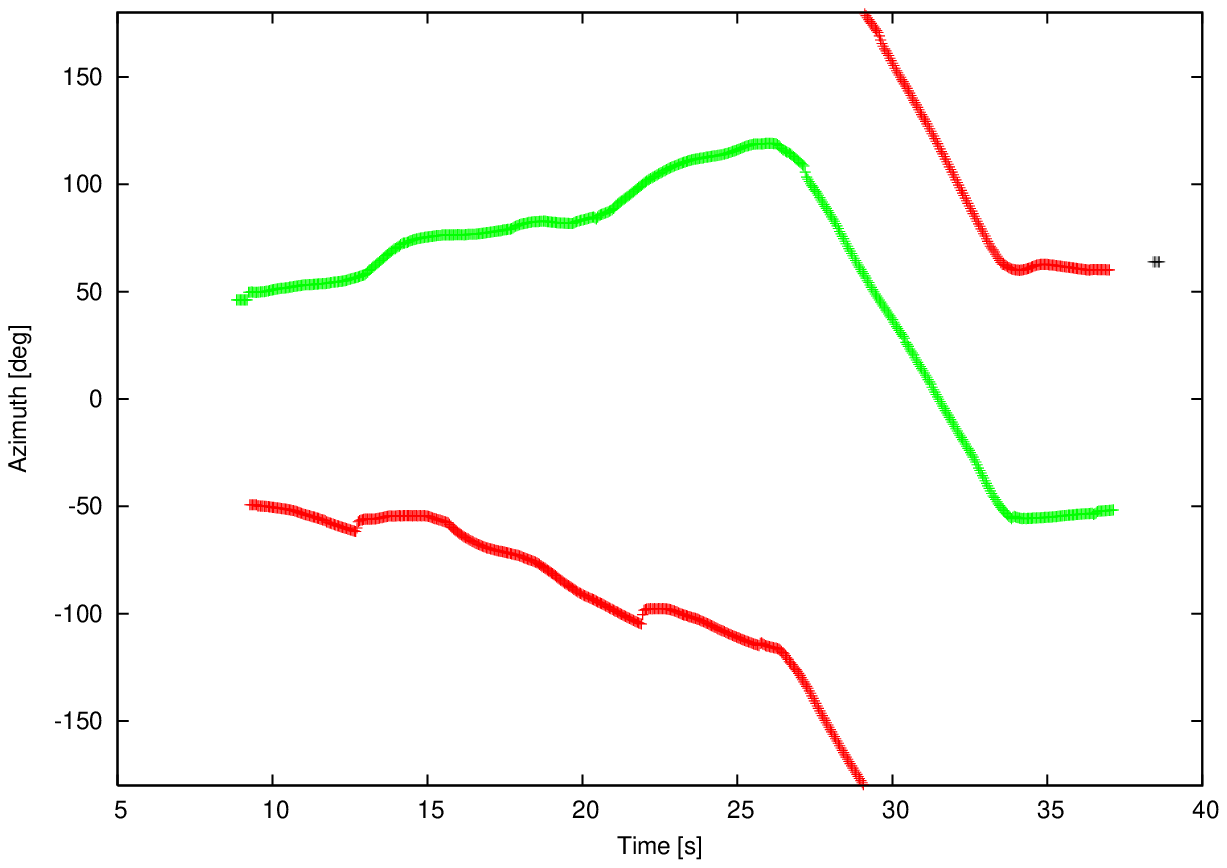}\includegraphics[width=2.5in]{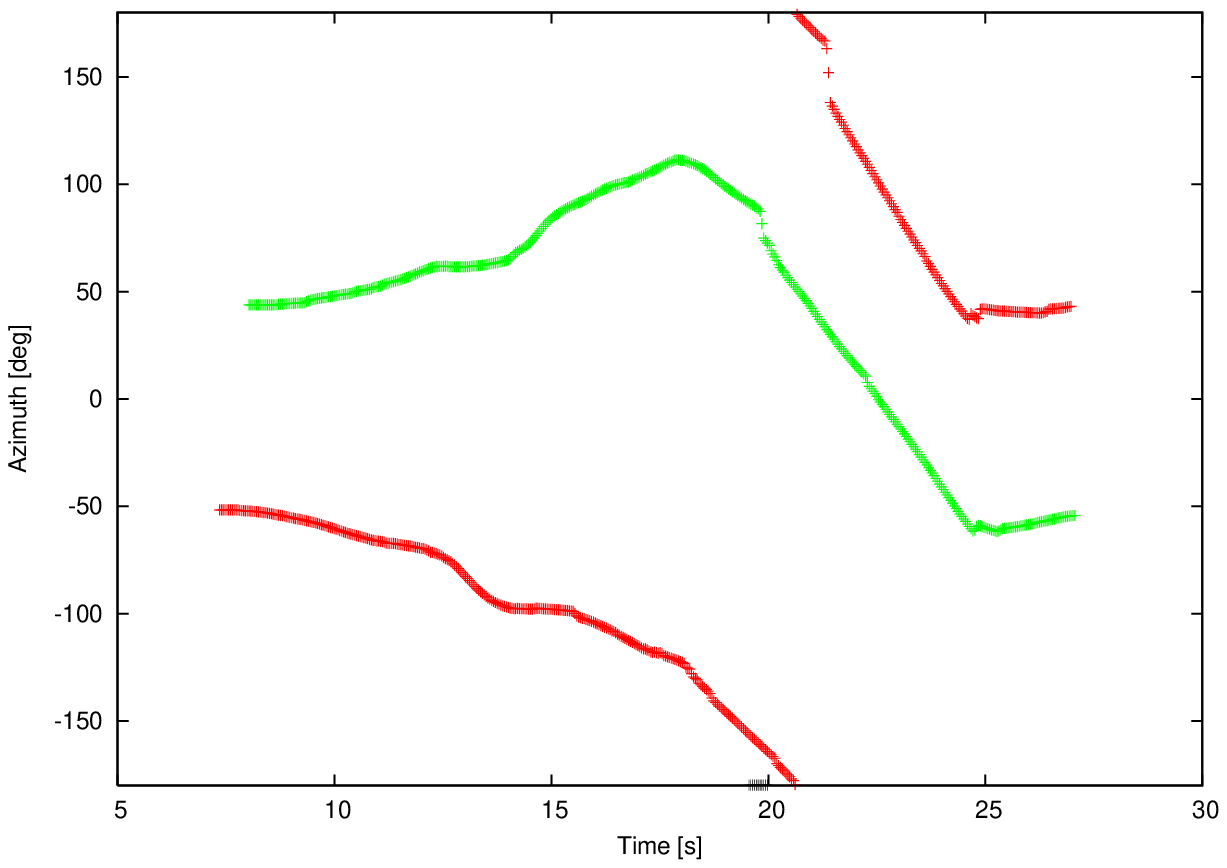}

\caption{Two stationary speakers with the robot moving. Left: E1, right: E2.
False detection shown in black. \label{cap:Two-speakers-robot-moving}}
\end{figure}

Tracking capabilities of our system are also evaluated in the context
where the robot is moving, as shown in the center of Figure \ref{cap:Source-trajectories}.
In this experiment, two people are talking continuously to the robot
as it is passing between them. The robot then makes a half-turn to
the left. Results are presented in Figure \ref{cap:Two-speakers-robot-moving}
for delayed estimation (500 ms). Once again, the estimated source
trajectories are consistent with the trajectories of the sources relative
to the robot for both environments. Only one false detection was present
(in E1, at $t=38\:\mathrm{s}$) for a short period of time.

\subsubsection{Sources with Intersecting Trajectories}

\begin{figure}[th]
\includegraphics[width=2.5in]{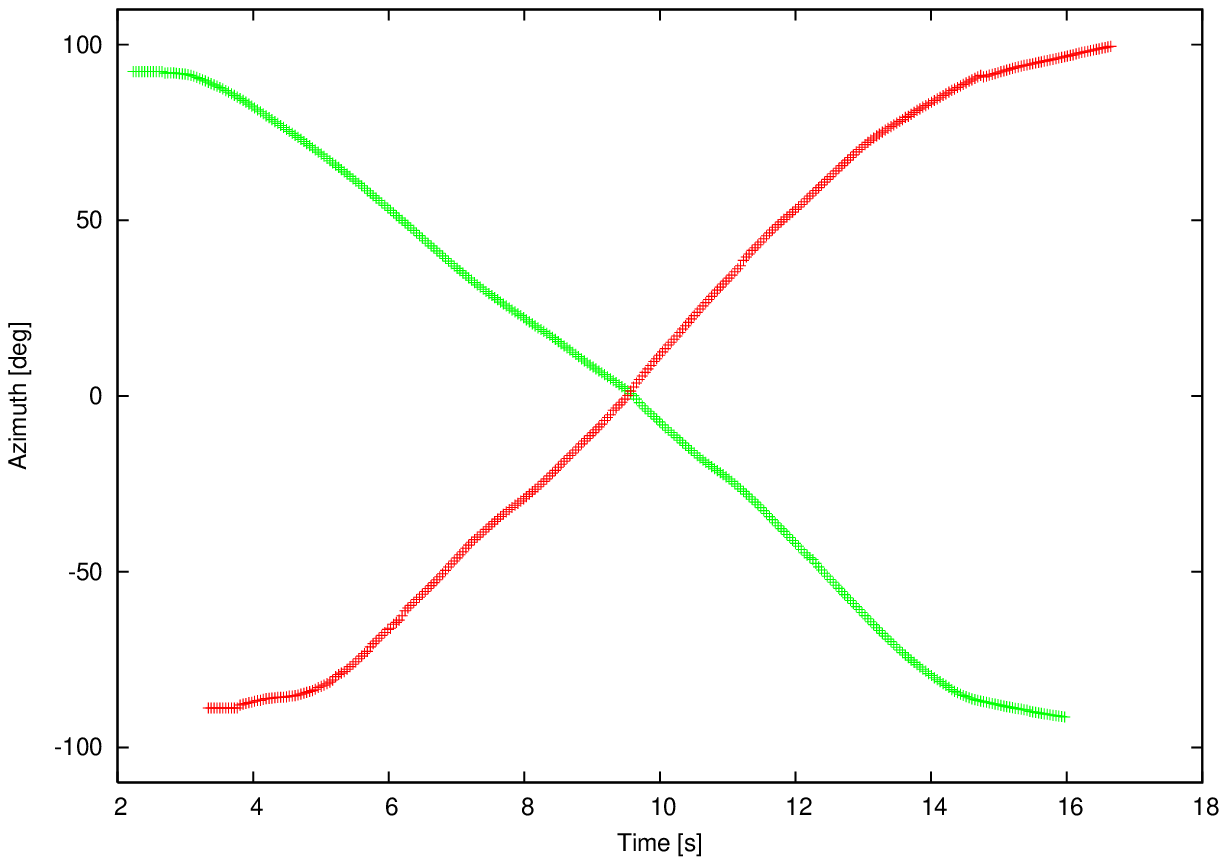}\includegraphics[width=2.5in]{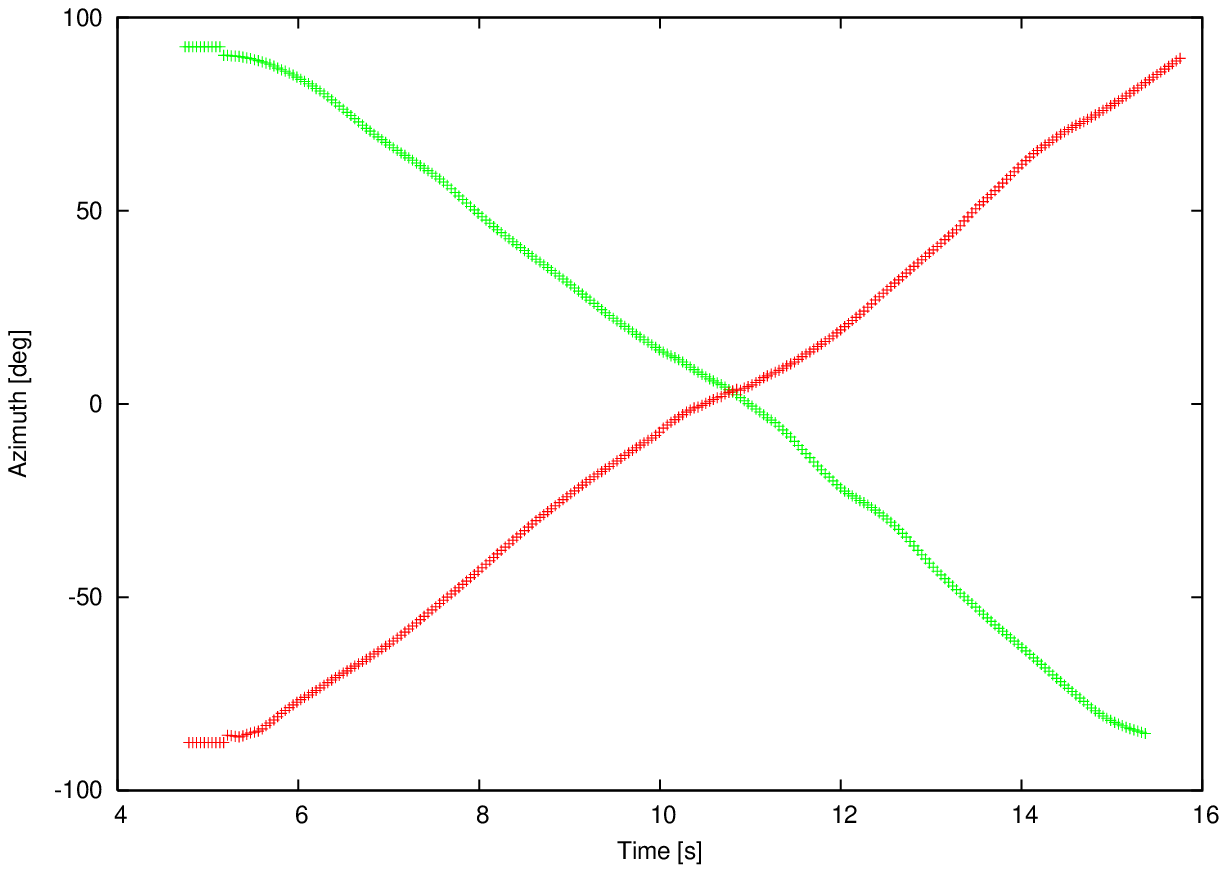}

\caption{Two speakers intersecting in front of the robot. Left: E1, right:
E2.\label{cap:Two-speakers-crossing}}
\end{figure}

In this experiment, two moving speakers are talking continuously to
the robot, as shown on the right of Figure \ref{cap:Source-trajectories}.
They start from each side of the robot, intersecting in front of the
robot before reaching the other side. Results in Figure \ref{cap:Two-speakers-crossing}
show that the particle filter is able to keep track of each source.
This result is possible because the prediction step imposes some inertia
to the sources and despite the fact that the steered beamformer typically
only {}``sees'' one source when the two sources are very close.

\subsubsection{Number of Microphones}

\begin{figure}[th]
\includegraphics[width=12cm,keepaspectratio]{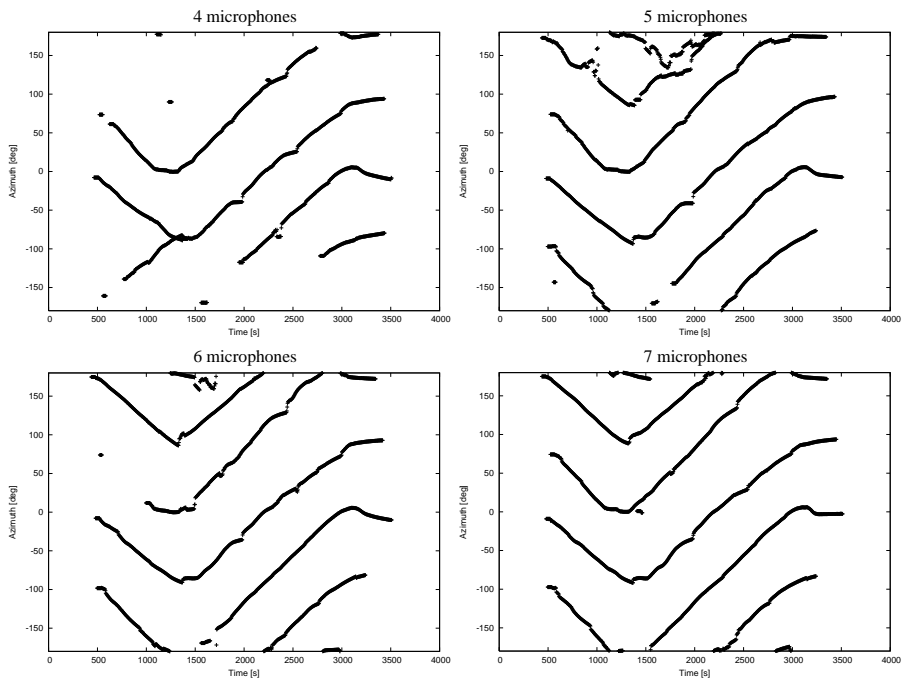}

\caption{Tracking of four sources using C2 in the E1 environment, using 4
to 7 microphones.\label{cap:Number-of-microphones}}
\end{figure}

These results evaluate how the number of microphones used affect the
system capabilities. To do so, we use the same recording as in \ref{sub:Moving-Sources}
for C2 in E1 with only a subset of the microphone signals to perform
localization. Since a minimum of four microphones are necessary for
localizing sounds without ambiguity, we evaluate the system for four
to seven microphones (selected arbitrarily as microphones number $1$
through $N$). Comparing results of Figure \ref{cap:Number-of-microphones}
to those obtained in Figure \ref{cap:Four-speakers-moving} for E1,
it can be observed that tracking capabilities degrade gracefully as
microphones are removed. While using seven microphones makes little
difference compared to the baseline of eight microphones, the system
is unable to reliably track more than two of the sources when only
four microphones are used. Although there is no theoretical relationship
between the number of microphones and the maximum number of sources
that can be tracked, this clearly shows the how redundancy added by
using more microphones can help in the context of sound source localization.

\subsection{Localization and Tracking for Robot Control}

This experiment is performed in real-time and consists of making the
robot follow the person speaking to it. At any time, only the source
present for the longest time is considered. When the source is detected
in front (withing 10 degrees) of the robot, it is made to go forward.
At the same time, regardless of the angle, the robot turns toward
the source in such a way as to keep the source in front. Using this
simple control system, it is possible to control the robot simply
by talking to it, even in noisy and reverberant environments. 

This has been tested by controlling the robot going from environment
E1 to environment E2, having to go through corridors and an elevator,
speaking to the robot with normal intensity at a distance ranging
from one meter to three meters. The system worked in real-time, providing
tracking data at a rate of 25 Hz (no additional delay on the estimator)
with the robot reaction time limited mainly by the inertia of the
robot. One problem we encountered during the experiment is that when
going through corridors, the robot would sometimes mistake reflections
on the walls for real sources. Fortunately, the fact that the robot
considers only the oldest source present reduces problems from both
reflections and noise sources.

\section{Conclusion\label{sec:Conclusion}}

Using an array of eight microphones, we have implemented a system
that is able to localize and track simultaneous moving sound sources
in the presence of noise and reverberation, at distances up to seven
meters. We have also demonstrated that the system is capable of controlling
in real-time the motion of a robot, using only the direction of sounds.
The tracking capabilities demonstrated result from combining our frequency-domain
steered beamformer with a particle filter tracking multiple sources.
Moreover, the original solution we found to the source-observation
assignment problem is also applicable to other multiple objects tracking
problems. Other novelties in this paper include the frequency-domain
implementation of our steered beamformer and the way we make it robust
to reverberation.

A robot using the proposed system has access to a rich, robust and
useful set of information derived from its acoustic environment. This
can certainly affect its ability of making autonomous decisions in
real life settings, and show higher intelligent behavior. Also, because
the system is able to localize multiple sound sources, it can be exploited
by a sound separation algorithm and enable speech recognition to be
performed. This will allow to identify the localized sound sources
so that additional relevant information can be obtained from the acoustic
environment.

\section*{Acknowledgment}

Jean-Marc Valin was supported by the National Science and Engineering
Research Council of Canada (NSERC) and the Quebec \emph{Fonds de recherche
sur la nature et les technologies} (FQRNT). Fran\c{c}ois Michaud
holds the Canada Research Chair (CRC) in Mobile Robotics and Autonomous
Intelligent Systems. This research is also supported financially by
the CRC Program and the Canadian Foundation for Innovation (CFI).
Special thanks to Brahim Hadjou for help formalizing the particle
filtering notation and to Dominic L\'{e}tourneau and Pierre Lepage
for help controlling the robot in real-time.

\bibliographystyle{elsart-num}
\bibliography{iros,BiblioAudible,localize}

\begin{thebibliography}{10}
\expandafter\ifx\csname url\endcsname\relax
  \def\url#1{\texttt{#1}}\fi
\expandafter\ifx\csname urlprefix\endcsname\relax\def\urlprefix{URL }\fi

\bibitem{Marschark}
M.~Marschark, Raising and Educating a Deaf Child, Oxford University Press,
  1998, http://www.rit.edu/~memrtl/course/interpreting/modules/modulelist.htm.

\bibitem{Hartmann1999}
W.~M. Hartmann, How we localize sounds, Physics Today (1999) 24--29.

\bibitem{ValinICRA2004}
J.-M. Valin, F.~Michaud, B.~Hadjou, J.~Rouat, Localization of simultaneous
  moving sound sources for mobile robot using a frequency-domain steered
  beamformer approach, in: Proceedings IEEE International Conference on
  Robotics and Automation, Vol.~1, 2004, pp. 1033--1038.

\bibitem{Mungamuru2004}
B.~Mungamuru, P.~Aarabi, Enhanced sound localization, IEEE Transactions on
  Systems, Man, and Cybernetics Part B 34~(3) (2004) 1526--1540.

\bibitem{nakadai-matsuura-okuno-kitano2003}
K.~Nakadai, D.~Matsuura, H.~G. Okuno, H.~Kitano, Applying scattering theory to
  robot audition system: Robust sound source localization and extraction, in:
  Proceedings IEEE/RSJ International Conference on Intelligent Robots and
  Systems, 2003, pp. 1147--1152.

\bibitem{Nakadai2000}
K.~Nakadai, T.~Lourens, H.~G. Okuno, H.~Kitano, Active audition for humanoid,
  in: Proceedings of the Seventeenth National Conference on Artificial
  Intelligence (AAAI), 2000, pp. 832--839.

\bibitem{Asano2001}
F.~Asano, M.~Goto, K.~Itou, H.~Asoh, Real-time source localization and
  separation system and its application to automatic speech recognition, in:
  Proc. EUROSPEECH, 2001, pp. 1013--1016.

\bibitem{ValinIROS2003}
J.-M. Valin, F.~Michaud, J.~Rouat, D.~L\'etourneau, Robust sound source
  localization using a microphone array on a mobile robot, in: Proceedings
  IEEE/RSJ International Conference on Intelligent Robots and Systems, 2003,
  pp. 1228--1233.

\bibitem{Kagami2004}
S.~Kagami, Y.~Tamai, H.~Mizoguchi, T.~Kanade, Microphone array for 2{D} sound
  localization and capture, in: Proceedings IEEE International Conference on
  Robotics and Automation, 2004, pp. 703--708.

\bibitem{Bechler2004}
D.~Bechler, M.~Schlosser, K.~Kroschel, System for robust 3{D} speaker tracking
  using microphone array measurements, in: Proceedings IEEE/RSJ International
  Conference on Intelligent Robots and Systems, 2004, pp. 2117--2122.

\bibitem{Arulampalam2002}
M.~S. Arulampalam, S.~Maskell, N.~Gordon, T.~Clapp, A tutorial on particle
  filters for online nonlinear/non-gaussian bayesian tracking, IEEE
  Transactions on Signal Processing 50~(2) (2002) 174--188.

\bibitem{Ward2002b}
D.~B. Ward, R.~C. Williamson, Particle filtering beamforming for acoustic
  source localization in a reverberant environment, in: Proceedings IEEE
  International Conference on Acoustics, Speech, and Signal Processing,
  Vol.~II, 2002, pp. 1777--1780.

\bibitem{Ward2003}
D.~B. Ward, E.~A. Lehmann, R.~C. Williamson, Particle filtering algorithms for
  tracking an acoustic source in a reverberant environment, IEEE Transactions
  on Speech and Audio Processing 11~(6).

\bibitem{Vermaak2001}
J.~Vermaak, A.~Blake, Nonlinear filtering for speaker tracking in noisy and
  reverberant environments, in: Proceedings IEEE International Conference on
  Acoustics, Speech, and Signal Processing, Vol.~5, 2001, pp. 3021--3024.

\bibitem{Asoh2004}
H.~Asoh, F.~Asano, K.~Yamamoto, T.~Yoshimura, Y.~Motomura, N.~Ichimura,
  I.~Hara, J.~Ogata, An application of a particle filter to bayesian multiple
  sound source tracking with audio and video information fusion, in:
  Proceedings of 7th International Conference on Information Fusion, 2004, pp.
  805--812.

\bibitem{Vermaak2003}
J.~Vermaak, A.~Doucet, P.~P\'{e}rez, Maintaining multi-modality through mixture
  tracking, in: Proceedings International Conference on Computer Vision (ICCV),
  2003, pp. 1950--1954.

\bibitem{MacCormick2000}
J.~MacCormick, A.~Blake, A probabilistic exclusion principle for tracking
  multiple objects, International Journal of Computer Vision 39~(1) (2000)
  57--71.

\bibitem{Hue2001}
C.~Hue, J.-P.~L. Cadre, P.~Perez, A particle filter to track multiple objects,
  in: Proceedings IEEE Workshop on Multi-Object Tracking, 2001, pp. 61--68.

\bibitem{Vermaak2005}
J.~Vermaak, S.~Godsill, P.~P\'{e}rez, Monte carlo filtering for multi-target
  tracking and data association, IEEE Transactions on Aerospace and Electronic
  Systems.

\bibitem{Duraiswami2001}
R.~Duraiswami, D.~Zotkin, L.~Davis, Active speech source localization by a dual
  coarse-to-fine search, in: Proceedings IEEE International Conference on
  Acoustics, Speech, and Signal Processing, 2001, pp. 3309--3312.

\bibitem{Omologo}
M.~Omologo, P.~Svaizer, Acoustic event localization using a crosspower-spectrum
  phase based technique, in: Proceedings IEEE International Conference on
  Acoustics, Speech, and Signal Processing, 1994, pp. {II--273}--{II--276}.

\bibitem{EphraimMalah1984}
Y.~Ephraim, D.~Malah, Speech enhancement using minimum mean-square error
  short-time spectral amplitude estimator, IEEE Transactions on Acoustics,
  Speech and Signal Processing ASSP-32~(6) (1984) 1109--1121.

\bibitem{CohenNonStat2001}
I.~Cohen, B.~Berdugo, Speech enhancement for non-stationary noise environments,
  Signal Processing 81~(2) (2001) 2403--2418.

\bibitem{Huang1997Echo}
J.~Huang, N.~Ohnishi, N.~Sugie, Sound localization in reverberant environment
  based on the model of the precedence effect, IEEE Transactions on
  Instrumentation and Measurement 46~(4) (1997) 842--846.

\bibitem{Huang1999Echo}
J.~Huang, N.~Ohnishi, X.~Guo, N.~Sugie, Echo avoidance in a computational model
  of the precedence effect, Speech Communication 27~(3-4) (1999) 223--233.

\bibitem{Giraldo97}
F.~Giraldo, Lagrange-galerkin methods on spherical geodesic grids, Journal of
  Computational Physics 136 (1997) 197--213.

\bibitem{Doucet2000}
A.~Doucet, S.~Godsill, C.~Andrieu, On sequential {M}onte {C}arlo sampling
  methods for bayesian filtering, Statistics and Computing 10 (2000) 197--208.

\bibitem{Hartmann1983}
W.~M. Hartmann, Localization of sounds in rooms, Journal of the Acoustical
  Society of America 74 (1983) 1380--1391.

\bibitem{Rakerd2004}
B.~Rakerd, W.~M. Hartmann, Localization of noise in a reverberant environment,
  in: Proceedings $18^{th}$ International Congress on Acoustics, 2004.

\end{thebibliography}

\end{document}